\pdfoutput=1

\documentclass[11pt]{article}

\usepackage[final]{acl}

\usepackage{times}
\usepackage{latexsym}

\usepackage[T1]{fontenc}

\usepackage[utf8]{inputenc}

\usepackage{microtype}

\usepackage{inconsolata}

\usepackage{graphicx}
\usepackage{enumitem}
\usepackage{listings}       
\usepackage{svg}
\usepackage{makecell}
\usepackage{tabularx}
\usepackage{longtable}
\usepackage{xcolor}         
\usepackage{pifont}
\usepackage{caption}        
\usepackage{tcolorbox}      
\usepackage{booktabs} 
\usepackage{multirow} 
\usepackage{colortbl}
\usepackage{amsmath}
\usepackage{float}
\usepackage{balance}
\definecolor{myyellow}{RGB}{255,171,64}
\definecolor{mylightyellow}{RGB}{252,229,205}

\lstdefinestyle{mystyle}{
    backgroundcolor=\color{mylightyellow},
    basicstyle=\ttfamily\footnotesize,
    breaklines=true,                 
    captionpos=b,                    
    numberstyle=\tiny\color{gray},
    stepnumber=1,
    numbersep=5pt,
    tabsize=2,
    frame=single,
    rulecolor=\color{myyellow},
    keywordstyle=\color{blue},
    stringstyle=\color{black},
    showstringspaces=false           %
}

\definecolor{myblue}{rgb}{0.2, 0.2, 1.0}

\tcbset{colback=gray!5, colframe=gray!80, fonttitle=\bfseries}

\usepackage[english, status=draft, margin=false, inline=true]{fixme}
\fxusetheme{color}
\FXRegisterAuthor{jf}{ajf}{\color{red}JF}
\FXRegisterAuthor{gs}{ags}{\color{blue}GS}

\lstdefinelanguage{JSON}{
    basicstyle=\ttfamily\footnotesize,
    keywordstyle=\bfseries\color{blue},
    stringstyle=\color{brown},
    commentstyle=\color{gray},
    showstringspaces=false,
    morestring=[b]",
    moredelim=[s][\color{blue}]{:}{\ },
    moredelim=[l][\color{magenta}]{,}
}

\title{Problem Solved? Information Extraction Design Space for Layout-Rich Documents using LLMs}

\author{
Gaye Colakoglu$^{1,2}$ \quad Gürkan Solmaz$^{2}$ \quad Jonathan Fürst$^{1}$ \\
$^{1}$Zurich University of Applied Sciences, Switzerland \\
$^{2}$NEC Laboratories Europe, Heidelberg, Germany \\
\texttt{colgay01@students.zhaw.ch, guerkan.solmaz@neclab.eu, jonathan.fuerst@zhaw.ch}
}

\newcommand{\rmspace}{\vspace{-0.0ex}}
\newcommand{\added}[1]{{\color{black}#1}}
\newcommand{\final}[1]{{\color{black}#1}}

\begin{document}
\maketitle
\begin{abstract}

This paper defines and explores the design space for information extraction (IE) from layout-rich documents using large language models (LLMs).
The three core challenges of layout-aware IE with LLMs are 1) data structuring, 2) model engagement, and 3) output refinement. Our study investigates the sub-problems and methods within these core challenges, such as input representation, chunking, prompting, selection of LLMs, and multimodal models. It examines the effect of different design choices through \textit{LayIE-LLM}, a new, open-source, layout-aware IE test suite, \added{benchmarking against traditional, fine-tuned IE models}. The results \added{on two IE datasets} show that LLMs require adjustment of the IE pipeline to achieve competitive performance: the optimized configuration found with LayIE-LLM achieves 13.3--37.5 F1 points more than a general-practice baseline configuration using the same LLM. To find a well-working configuration, we develop a one-factor-at-a-time (OFAT) method that achieves near-optimal results. Our method is only \added{0.8--1.8} points lower than the best full factorial exploration with a fraction ($\sim$2.8\%) of the required computation.
Overall, we demonstrate that, if well-configured, general-purpose LLMs match the performance of specialized models, providing a cost-effective, finetuning-free alternative.
Our test-suite is available at \url{https://github.com/gayecolakoglu/LayIE-LLM}.

\end{abstract}

\section{Introduction}
Information extraction (IE) extracts structured data from unstructured documents, including layout-rich documents (LRDs) as reports and presentations that mix visual and textual elements~\cite{park2019cord, Wang_2023, zmigrod2024buddiebusinessdocumentdataset}. LRDs challenge traditional natural language processing (NLP) techniques, designed for plain texts~\cite{cui2021document, tang2023unifying}.

\begin{figure}[ht]
    \centering    
    \includegraphics[width=0.48\textwidth,trim={0 0 0 0.09cm},clip]{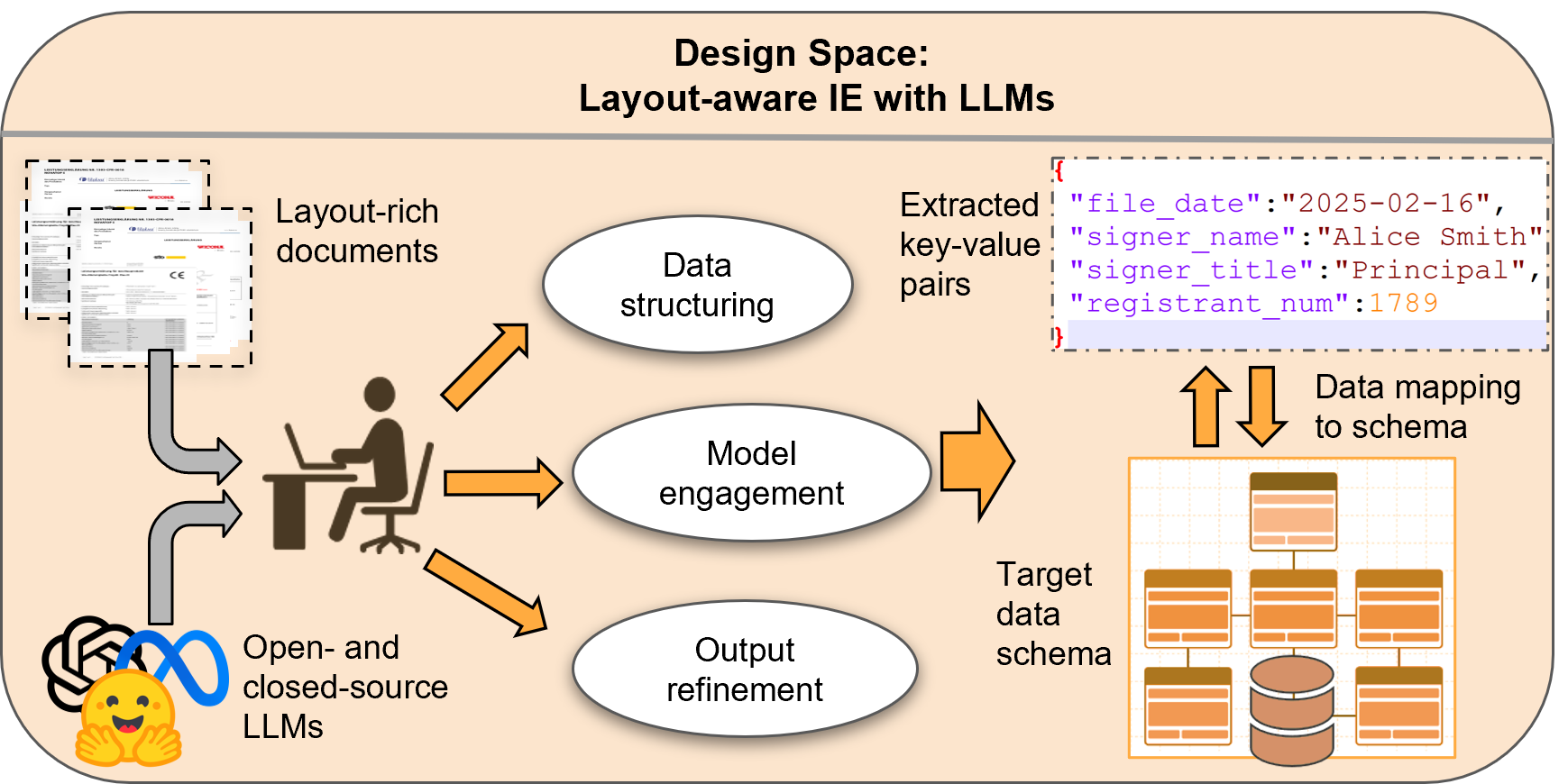}
    \caption{Design space for IE from layout-rich documents using LLMs. The goal is to extract information relevant to the target data schema with correct mapping.}
    \label{fig:design_space}
    \rmspace
\end{figure}

Recent layout-aware models at the intersection of NLP and Computer Vision (CV) address this gap by including visual and structural features to improve IE from LRDs (e.g., LayoutLMv1-v3~\cite{xu2020layoutlm, xu2020layoutlmv2, huang2022layoutlmv3}, \added{ERNIE-Layout~\cite{peng2022ernie}}, GraphDoc~\cite{zhang2022multimodal}, DocFormer~\cite{appalaraju2021docformer}).
\final{While early layout-aware models required substantial dataset-specific fine-tuning, recent generative document understanding models~\cite{powalski2021going} have demonstrated strong zero- and few-shot capabilities on unseen datasets and tasks.}

While LLMs exhibit significant potential, there exist open questions regarding their adoption for IE from LRDs. In what form should the document content be provided to the LLM? Which methods are most effective for in-context learning (ICL) and instruction tuning? \added{Do large multimodal LLMs as GPT-4o~\cite{openai2024gpt4o} and Qwen 2.5-vision~\cite{bai2025qwen2} offer advantages over traditional Optical Character Recognition (OCR) and fine-tuned layout-aware models, and how does their performance vary across LLMs of different scales and sources?} How coherent is the LLM output, and what post-processing is needed?

\added{Studies have individually addressed some of the aspects~\cite{wang2023docllm, luo2024layoutllm, blau2024context}}. However, there is a lack of a unified framework that comprehensively explores and compares these components. This paper fills this gap by systematically investigating the design space for LLM-based IE from LRDs. We focus on evaluating a range of preprocessing, chunking, prompting, and post-processing techniques, alongside various ICL strategies, within a unified, reproducible framework. \added{Additionally, we conduct an extensive comparison of these components, examining multimodal LLMs, traditional models, open- and closed-source LLMs, and fine-tuned layout-aware models. \textit{To the best of our knowledge, no existing research has comprehensively compared the combined impact of these pipeline parameters across such diverse model categories.}}

Our results reveal multiple insights. First, current general LLMs can compete with \added{traditional fine-tuned models such as LayoutLMv3 and ERNIE-Layout}, without expensive labeling. Second, rather than fine-tuning on data, LLMs require adjustment of the IE pipeline to achieve competitive performance--\added{the performance gap between a general-practice baseline and the configuration tuned using our lightweight OFAT method on two datasets is 13.3 F1 points for VRDU~\cite{Wang_2023}, and 37.5 for FUNSD~\cite{jaume2019funsd}, only 1.8 points and 0.8 points lower than the best possible Brute-Force configuration respectively.} Third, while purely text-based LLMs achieve competitive performance with our method, multimodal LLMs, those that directly integrate textual and visual features, still outperform them. However, this advantage comes with increased token usage, higher API costs, and diminished transparency and control over individual steps. In summary, our contributions are as follows.

\begin{itemize}[leftmargin=*,noitemsep,topsep=0pt]

\item We introduce the \textbf{Design Space of IE from LRDs} using LLMs, addressing three core challenges: 1) Data structuring, 2) Model engagement, and 3) Output refinement (Sec.~\ref{sec:design_space_IE}).

\item We develop LayIE-LLM, a \textbf{layout-aware IE test suite} to analyze the impact of: OCR, text-based inputs, chunk size, few-shot and CoT prompting, LLM models, decoding strategies, schema mapping, data cleaning, and evaluation techniques using exact, substring, and fuzzy matching  (Sec.~\ref{sec:test_suite_IE_layout}).

\item \added{We conduct evaluation of \textbf{GPT-4o}, \textbf{GPT-3.5}, and \textbf{LLaMA3}, as well as the vision-enabled \textbf{GPT-4o \final{and Qwen2.5-vision}}, and compare them with traditional, fine-tuned layout-aware models such as \textbf{LayoutLMv3} and \textbf{ERNIE-Layout}, highlighting trade-offs between performance and resource efficiency (Sec.~\ref{sec:experimental_evaluation}).}

\item We open-source LayIE-LLM and all our experimental results, enabling reproducibility and further experimentation by the community: \url{https://github.com/gayecolakoglu/LayIE-LLM} 

\end{itemize}

\section{Design Space of IE from LRDs with LLMs}
\label{sec:design_space_IE}

\paragraph{Task Definition.}IE from LRDs involves identifying and extracting information from documents where textual content is intertwined with complex visual layouts and mapping them into structured information instances such that

\begin{equation}
\text{IE}: (D, S) \rightarrow E
\end{equation}
\begin{itemize}[leftmargin=*,noitemsep,topsep=0pt]
    \item \textbf{D} represents the set of LRDs, each with content and layout information.
    \item \textbf{S} is the target schema that defines the set of slots to be filled.
    Each slot is defined by an attribute (key) $a_i$ and its corresponding data type (domain) $T_i$, such that $S = \{(a_1, T_1), (a_2, T_2), ..., (a_k, T_k)\}$.
    
    \item Finally, \textbf{E} represents the set of extracted information instances, where each instance is a set of slot-value pairs derived from a document in \textbf{D}, leveraging both content and layout to determine the correct values for the slots in \textbf{S}. Each value in an instance must conform to the data type $T$, specified in the schema for that attribute.
    
\end{itemize}

\subsection{Using LLMs for IE}
LLM-based IE systems have to tackle three challenge areas: \textit{Data Structuring}, \textit{Model Engagement}, and \textit{Output Refinement}.

\noindent \textbf{Data Structuring.} For multimodal LLMs, LRDs can be directly given as input. On the other hand, for purely text-based LLMs, the input documents must be transformed into textual representations. This involves converting documents into machine-readable formats using OCR systems to extract features such as text, bounding boxes, and visual elements~\cite{mieskes-schmunk-2019-ocr, smith2007overview}. 
Alternatively, a formatting language such as Markdown can be employed to represent the document's layout, allowing the LLM to understand the structural context of the text better.
The impact of OCR quality on IE performance has been documented~\cite{bhadauria-etal-2024-effects}, and structured formats tend to yield better results~\cite{bai-etal-2024-schema}.
To process larger documents efficiently, they are often divided into smaller, manageable chunks based on page boundaries, sections, or semantic units~\cite{liu2024lost}.
\colorbox{blue!10}{\parbox{\columnwidth}{\emph{Markdown as an input format compared to raw OCR outputs remains underexplored, representing a potential research gap in IE system development.}}}

\noindent \textbf{Model Engagement.} Once preprocessed, the document is fed to an LLM for IE. Ensuring alignment between the extracted text and layout information is crucial for accurate representation~\cite{xu2020layoutlmv2, appalaraju2021docformer}. Prompt-driven extraction leverages general-purpose models, by using tailored prompts to guide the extraction process~\cite{brown2020language, radford2021learning, zhou2022least}. As such, models must be explicitly instructed to extract information, typically with an accompanying schema. This step can involve more advanced IT and ICL techniques (few-shot, CoT).
\colorbox{blue!10}{\parbox{\columnwidth}{\emph{The influence of prompting techniques in interaction with various stages of the IE pipeline to enhance performance and robustness remains a research gap that requires further investigation.}}}

\noindent \textbf{Output Refinement.} After inference, the extracted information undergoes post-processing to ensure accuracy and conformity with Schema $S$. This step involves refining and validating the outputs generated by the LLM through tasks such as mapping extracted entities $E$ to their original document positions, merging overlapping or fragmented predictions, and resolving ambiguities in the results~\cite{xu2020layoutlm}. Post-processing for entity extraction has been explored in prior studies~\cite{j-wang-etal-2022-globalpointer, tamayo-etal-2022-nlp}, and rule-based entity alignment has demonstrated notable improvements in accuracy~\cite{luo2024asgeaexploitinglogicrules}.
\colorbox{blue!10}{\parbox{\columnwidth}{\emph{However, to our knowledge, there has been no analysis of post-processing techniques specifically tailored to LRDs in conjunction with LLMs.
}}}

\section{LayIE-LLM: Test-Suite for IE from LRDs with LLMs}
\label{sec:test_suite_IE_layout}

\begin{figure*}[ht]
    \centering
    \includegraphics[width=1.0\textwidth]{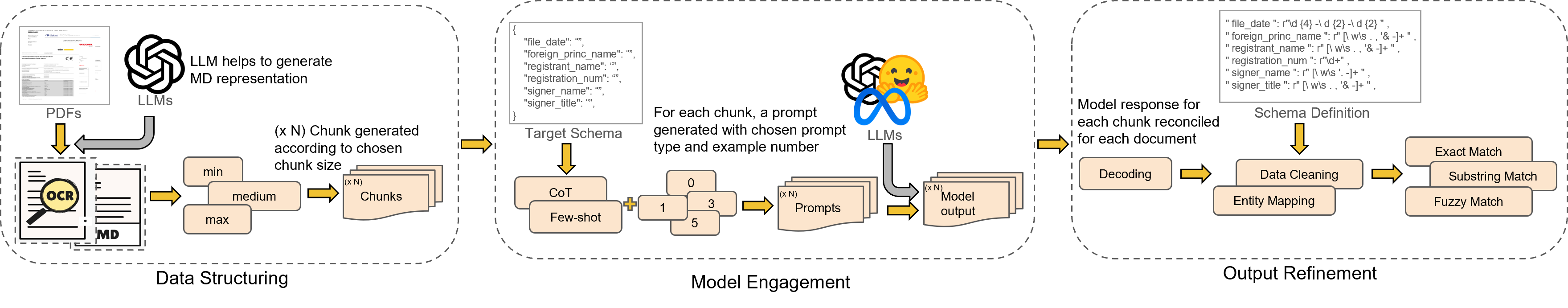}
    \caption{LayIE-LLM test suite for extracting information from LRDs using LLMs in six stages.
    The process begins with OCR-based text extraction and Markdown conversion with LLM assistance, followed by chunking to manage token limits, experimenting with different chunk sizes. Each chunk is processed into a prompt using Few-shot and CoT strategies with varying example counts. Prompts include a document example with key-value pairs, plus a new document and task. LLMs generate structured JSON outputs, which are decoded and reconciled. Post-processing includes data cleaning and entity mapping, followed by evaluation using two methods.}
    \label{fig:pipeline}
    \rmspace
\end{figure*}

We implement \href{https://github.com/gayecolakoglu/LayIE-LLM}{LayIE-LLM} , a comprehensive test suite to evaluate IE tasks from LRDs. The pipeline, depicted in Figure~\ref{fig:pipeline}, systematically transforms raw inputs into structured outputs through multiple stages, enabling a thorough assessment of the effectiveness of various design decisions.

\subsection{Data Structuring}
PDF documents include both textual content and layout features. The process involves two conversions: 
1) Extracting textual content and layout information using OCR data, and 2) creating a markdown representation.

\noindent \textbf{Chunking.}
We employ three chunk sizes: (1) max: 4096 tokens, (2) medium: 2048 tokens, and (3) small: 1024 tokens. Documents are segmented into $N$ chunks based on their length and the selected chunk size. Each chunk is constructed by sequentially accumulating whole words up to the token limit, thereby preserving word boundaries and ensuring a non-overlapping structure. Layout information is retained by associating each text segment with normalized and quantized spatial coordinates, which preserve the structural context of the original document.
\subsection{Model Engagement}
Model engagement involves constructing input to the LLM using three primary components: (1) a task instruction that defines the IE task, (2) the target schema $S$, and (3) the document chunk.
We adhere to best practices from NLP for prompt structure and IE task instruction. The schema $S$ is implemented as a dictionary of key-value pairs, where values specify the format of the corresponding attribute using regex expressions in Listing~\ref{lst:schema}.

\begin{lstlisting}[language=Python, caption={Schema Definition}, label={lst:schema}]
"file_date": r"\d{4}-\d{2}-\d{2}",
"foreign_princ_name": r"[\w\s.,'&-]+",
"registrant_name": r"[\w\s.,'&-]+",
"registration_num": r"\d+",
"signer_name": r"[\w\s'.-]+",
"signer_title": r"[\w\s.,'&-]+",
\end{lstlisting}
We also implement two ICL strategies: (1) Few-Shot and (2) Chain-of-Thought (CoT)
(see Appendix~\ref{appendix:prompt_generation_details} for more details).
For $N$ prompts, the LLM performs \emph{N completions}, with one completion per prompt. The outputs are collected and stored as raw predictions, ready for Output Refinement.

\subsection{Output Refinement}
\label{subsec:post_processing}
We refine the raw outputs from the LLM to ensure alignment with the target schema, addressing challenges related to prediction variability, schema definition differences, and data formatting inconsistencies (e.g., varying date formats). We implement three techniques inspired by related work in data integration~\cite{dong2013big}: \textit{Decoding}, \textit{Schema Mapping}, and \textit{Data Cleaning}. This results in three sets of predictions: initial predictions, mapped predictions, and cleaned predictions.

\noindent \textbf{Decoding.} The decoding step parses each LLM completion as a JSON object, discarding any that fail to parse. The process then consolidates predictions for each document by reconciling outputs generated across individual pages and chunks. With $N$ completions for $N$ prompts, corresponding to $N$ chunks, the model generates multiple predictions for a single document. Reconciliation ensures a unified document-level output by deduplicating nested predictions and aggregating unique values. If multiple unique values exist for a single entity, they are retained to preserve variability. The outcome of this step is referred to as  \emph{initial predictions}.

\noindent \textbf{Schema Mapping.} LLMs are expected to return only keys $\{a_1, a_2, ..., a_k\}$ specified in the target schema $S$. 
However, they may occasionally fail to return the keys as expected. E.g., `file date'' is returned instead of ``file\_date''. Such LLM ``overcorrection'' can hinder strict schema conformance.
As a countermeasure, we implement a post-processing step that maps the predicted keys to align with the target schema.
Our mapping step integrates multiple weak-supervision signals, such as \textit{exact matching, partial matching,} and \textit{synonym-based logic}, inspired by the recent techniques for ontology alignment~\cite{furst2023versamatch}.
The outcome of this step is the \emph{mapped predictions}, where entity keys $\{a_1, a_2, ..., a_k\}$ are standardized and fully aligned with the schema $S$.

\noindent \textbf{Data Cleaning.} A common issue concerns the format of the values $T_k$ of predicted key-value pairs $\{(a_1, T_1), (a_2, T_2), ..., (a_k, T_k)\}$. We must standardize formats, such as dates and names, to align with the target schema $S$.
One source of error is LLM hallucinations~\cite{ji2023survey, huang2023survey, xu2024hallucination}, while another problem is that information is often not aligned to a common format inside the source data. For instance, two documents might use two different formats for dates (``April 1992'' vs ``1992-04-01''). Additional issues include capitalization, redundant whitespace, or special characters. We utilize the regex-defined data types in our schema to automatically apply data cleaning functions. 
The outcome of this step is the \emph{cleaned predictions}, representing the final fully normalized outputs.

\noindent \textbf{Evaluation Techniques}.
\label{subsec:evaluation_techniques} 
Evaluating IE for LRDs requires comparing the extracted data against an annotated test dataset. We implement three metrics for this evaluation: \textit{exact match, substring match}, and \textit{fuzzy match}.

\begin{itemize}[leftmargin=*,noitemsep,topsep=0pt]
\item \textbf{Exact Match} searches for perfect alignment between predicted and ground truth values. A match is valid only when the values are identical. This strict approach is ideal for extracting specific, unambiguous entities like dates or numerical identifiers.

\item \textbf{Substring Match} checks whether ground truth values are fully contained within the predicted values as complete substrings, without being split or partially matched. It ensures all ground truth values appear in their entirety within predictions, making it effective for tasks such as extracting full names or addresses, where additional contextual details (e.g., titles like Mr. and Mrs.) may be included in the predictions without making the extraction incorrect.

\item \textbf{Fuzzy Match} uses similarity metrics \final{(fuzz.ratio)} for approximate matches. A match is valid if the highest similarity ratio exceeds a predefined threshold (default: 0.8), \final{creating binary decisions unlike soft similarity metrics~\cite{biten2019scene}}. This method is well-suited for scenarios with minor variations caused by OCR errors or formatting discrepancies.
\end{itemize}

\section{Experimental Evaluation}
\label{sec:experimental_evaluation}

\subsection{Experimental Design}
\noindent \textbf{Methodology.} The goal of our experimental setup is to analyze how different parameters in the pipeline (Figure~\ref{fig:pipeline}) influence the performance of IE from LRDs using LLMs. To examine the design dimensions, we begin with a baseline configuration and systematically vary one factor at a time, following a one-factor-at-a-time (OFAT) methodology. This approach allows us to isolate and understand the impact of each parameter change on the IE performance.

\noindent Our intuition is that by independently aggregating the insights gained from each design dimension, one can develop a deeper understanding of the design space and potentially identify an effective overall configuration for IE, without the need for an exhaustive factorial exploration. To validate this approach, we compare the OFAT method’s results with those obtained via a brute-force strategy involving 432 experiments ($2\times3\times2\times4\times3\times3=432$, see Table~\ref{tab:all_conf}).

\noindent \textbf{Dataset and LLMs.}
We use the Visually Rich Document Understanding (VRDU) dataset~\cite{Wang_2023}, which includes two benchmarks with high-quality OCR for assessing data efficiency, due to its inclusion of three visually diverse template types – Amendment, Short-Form, and Dissemination – representing varied and complex layouts, as well as generalization tasks: \textit{Single Template Learning (STL), Unseen Template Learning (UTL)}, and \textit{Mixed Template Learning (MTL)}. For further details on how this dataset is tailored for our experiments and diverse models, please refer to~\ref{appendix:dataset_details}.

\noindent \added{Additionally, to further prove the generalizability of our approach, we also conducted the same set of experiments using the FUNSD dataset~\cite{jaume2019funsd}, which directly provides key-value pairs in a question-answering format without the complexity of OCR processing. Due to page limitations and for the sake of clarity, the experimental results presented in this section are based on the VRDU dataset. The corresponding experimental results for the FUNSD dataset are provided in~\ref{appendix:funsd_results}.}

We evaluate GPT-3.5, GPT-4o, and LLaMA3-70B for text-only structured data extraction from LRDs. Furthermore, we compare their results with those of GPT-4, LayoutLMv3, and ERNIE-Layout to assess the performance gap between multimodal LLMs and domain-specific fine-tuned models.
\begin{table}[hbtp]
\centering
\footnotesize
\caption{Overall configuration parameters. Baseline configuration is highlighted with \colorbox{blue!10}{light blue}.}
\label{tab:all_conf}
\begin{tabular}{p{3.2cm}p{3.6cm}}
\toprule  
\textbf{Parameter}       & \textbf{Values}                         \\ \midrule
Input Type              & \colorbox{blue!10}{OCR}, Markdown          \\ 
Chunk Size Category     & Small, \colorbox{blue!10}{Medium}, Max     \\ 
Prompt Type             & \colorbox{blue!10}{Few-Shot}, CoT         \\ 
Example Number          & \colorbox{blue!10}{0}, 1, 3, 5             \\ 
Post-processing Strategy & \shortstack[l]{\colorbox{blue!10}{Initial}, Mapped, Cleaned} \\
Evaluation Technique    & \shortstack[l]{\colorbox{blue!10}{Exact}, Substring, Fuzzy} \\
 \bottomrule
\end{tabular}
\rmspace
\end{table}

\noindent \textbf{The Baseline Configuration.}
The baseline configuration is outlined in Table~\ref{tab:all_conf}, where the configuration is selected based on best practices, such as in \cite{perot-etal-2024-lmdx}
for the following reasons: (1) \textbf{OCR} reflects real-world scenarios for digitized LRDs. (2) \textbf{Medium chunk size} balances efficiency and context preservation, addressing token limits in LLMs. (3) \textbf{Few-shot prompting} combines pre-trained knowledge with minimal task-specific guidance. (4) Using \textbf{zero examples} provides a clear benchmark for assessing the model's raw performance. (5) \textbf{Initial predictions} are retained to evaluate models' raw output without modifications, ensuring a direct assessment of their capabilities. (6) Finally, \textbf{exact match} provides a stringent measure of correctness, offering a reliable baseline for comparison across configurations.

\subsection{The Input Dimension}

We substitute OCR input with Markdown and evaluate the resulting performance in both STL and UTL scenarios. The performance differences between OCR and Markdown inputs vary by model and context, showing no consistent trend favoring either input type, as shown in Table~\ref{tab:input_type}.

\begin{table}[hbtp]
    \centering %
    \scriptsize %
    \caption{Performance results for different LLMs across \textbf{STL} and \textbf{UTL} levels with different input types. Baseline configuration in \colorbox{blue!10}{light blue}.} 
    \label{tab:input_type}  
    \setlength{\tabcolsep}{4pt}
    \begin{tabular}{l l l l l}
        \toprule  
        \multirow{2}{*}{\textbf{Models}} & \multirow{2}{*}{\textbf{Level}} & \multicolumn{2}{c}{\textbf{Exact Match (F1)}} \\  
        \cmidrule(lr){3-4}
        & & \textbf{OCR} & \textbf{Markdown} \\
        \midrule
        \multirow{2}{*}{GPT-3.5} & STL & \colorbox{blue!10}{0.650} & 0.647 \textbf{{\color{gray}\fontsize{5.5}{8.4}\selectfont(-0.003) }} \\  
                                 & UTL & \colorbox{blue!10}{0.645} & 0.657 \textbf{{\color{gray}\fontsize{5.5}{8.4}\selectfont(+0.012) }} \\ 
        \cmidrule(lr){2-4}
        \multirow{2}{*}{GPT-4o} & STL & \colorbox{blue!10}{0.670} & 0.633 \textbf{{\color{gray}\fontsize{5.5}{8.4}\selectfont(-0.037) }} \\  
                                 & UTL & \colorbox{blue!10}{0.659} & 0.633 \textbf{{\color{gray}\fontsize{5.5}{8.4}\selectfont(-0.026) }} \\  
        \cmidrule(lr){2-4}
        \multirow{2}{*}{LLaMA3} & STL & \colorbox{blue!10}{0.640} & 0.657 \textbf{{\color{gray}\fontsize{5.5}{8.4}\selectfont(+0.017) }} \\ 
                                    & UTL & \colorbox{blue!10}{0.640} & 0.662 \textbf{{\color{gray}\fontsize{5.5}{8.4}\selectfont(+0.022) }} \\ 
    \bottomrule  
    Avg \textbf{{\color{gray}\fontsize{5.5}{8.4}\selectfont(\textpm stdev.) }} & & \colorbox{blue!10}{0.650 \textbf{{\color{gray}\fontsize{5.5}{8.4}\selectfont(\textpm 0.011) }}} & 0.648 \textbf{{\color{gray}\fontsize{5.5}{8.4}\selectfont(\textpm 0.012) }}
    \end{tabular}
    \rmspace
\end{table}

\noindent OCR input serves as a stable baseline for IE tasks, delivering consistent performance across models. GPT-4o has noticeable performance drops with Markdown input, indicating its reliance on OCR for optimal results. In contrast, Markdown marginally improves performance for LLaMA3-70B at both STL and UTL scenarios, suggesting its potential benefits from the additional structure or semantic cues. GPT-3.5 demonstrates robustness to changes in input type, with only slight fluctuations in performance. On average, OCR marginally outperforms Markdown (0.650 vs. 0.648), but the differences are minor, with standard deviations indicating similar stability (see Appendix~\ref{appendix:observations} for detailed insights).

\subsection{The Chunk Dimension}
\begin{table}[hbtp]
    \small
    \caption{Performance results for different LLMs across \textbf{STL} and \textbf{UTL} levels with different chunk size categories. Baseline configuration in \colorbox{blue!10}{light blue}.} 
    \label{tab:chunk_size}
    \resizebox{\linewidth}{!}{
    \setlength{\tabcolsep}{4pt}
    \begin{tabular}{l c c c c}
    \toprule
    \multirow{2}{*}{\textbf{Models}} & \multirow{2}{*}{\textbf{Level}} & \multicolumn{3}{c}{\textbf{Exact Match (F1)}} \\
    \cmidrule(lr){3-5}
    & & \textbf{Small\textbf{{\color{gray}\fontsize{7}{8.4}\selectfont($\leq 1024$)}}} & \textbf{Medium\textbf{{\color{gray}\fontsize{7}{8.4}\selectfont($\leq 2048$)}}} & \textbf{Max\textbf{{\color{gray}\fontsize{7}{8.4}\selectfont($\leq 4096$)}}} \\
    \midrule
    \multirow{2}{*}{GPT-3.5} & STL & 0.562\textbf{{\color{gray}\fontsize{5.5}{8.4}\selectfont(-0.088) }} & \colorbox{blue!10}{0.650} & 0.645\textbf{{\color{gray}\fontsize{5.5}{8.4}\selectfont(-0.005) }} \\
             & UTL & 0.561\textbf{{\color{gray}\fontsize{5.5}{8.4}\selectfont(-0.084) }} & \colorbox{blue!10}{0.645} & 0.644\textbf{{\color{gray}\fontsize{5.5}{8.4}\selectfont(-0.001) }} \\
    \cmidrule(lr){2-5}
    \multirow{2}{*}{GPT-4o} & STL & 0.602\textbf{{\color{gray}\fontsize{5.5}{8.4}\selectfont(-0.068) }} & \colorbox{blue!10}{0.670} & 0.674\textbf{{\color{gray}\fontsize{5.5}{8.4}\selectfont(+0.004) }} \\
             & UTL & 0.600\textbf{{\color{gray}\fontsize{5.5}{8.4}\selectfont(-0.059) }} & \colorbox{blue!10}{0.659} & 0.657\textbf{{\color{gray}\fontsize{5.5}{8.4}\selectfont(-0.002) }} \\
    \cmidrule(lr){2-5}
    \multirow{2}{*}{LLaMA3} & STL & 0.615\textbf{{\color{gray}\fontsize{5.5}{8.4}\selectfont(-0.025) }} & \colorbox{blue!10}{0.640} & 0.647\textbf{{\color{gray}\fontsize{5.5}{8.4}\selectfont(+0.007) }} \\
               & UTL & 0.608\textbf{{\color{gray}\fontsize{5.5}{8.4}\selectfont(-0.032) }} & \colorbox{blue!10}{0.640} & 0.644\textbf{{\color{gray}\fontsize{5.5}{8.4}\selectfont(+0.004) }} \\
    \bottomrule
    Avg\textbf{{\color{gray}\fontsize{5.5}{8.4}\selectfont(\textpm stdev.) }} & & 0.591\textbf{{\color{gray}\fontsize{5.5}{8.4}\selectfont(\textpm 0.023) }} & \colorbox{blue!10}{0.650\textbf{{\color{gray}\fontsize{5.5}{8.4}\selectfont(\textpm 0.011) }}} & 0.651\textbf{{\color{gray}\fontsize{5.5}{8.4}\selectfont(\textpm 0.011) }}
    \end{tabular}}
    \rmspace
\end{table}
To evaluate the impact of chunk size, we vary it while keeping all other parameters constant.
Table~\ref{tab:chunk_size} demonstrates how chunk size affects performance across STL and UTL levels.

Medium and max chunk sizes provide the most consistent and stable results across models, with an average F1 score of 0.650 (±0.011) and 0.651 (±0.011), respectively.
Due to insufficient context, small chunk sizes result in significant performance drops, particularly for GPT-3.5 and GPT-4o.
\emph{These findings suggest that max chunk size is optimal, but medium can be a good option for LLMs with limited context lengths}.

\subsection{The Prompt Dimension}
\begin{table}[hbtp!]
    \centering %
    \scriptsize %
    \caption{Different LLMs across \textbf{STL} and \textbf{UTL} levels with different prompt types and example numbers. Baseline Configuration is highlighted in \colorbox{blue!10}{light blue}.}
\label{tab:prompt_type_with_examples}  
    \setlength{\tabcolsep}{1pt}
    \begin{tabular}{l c c c c c}
    \toprule  
    \multirow{2}{*}{\textbf{Models}} & \multirow{2}{*}{\textbf{Level}} & 
    \multicolumn{4}{c}{\textbf{Exact Match (F1)}} \\  
    \cmidrule(lr){3-6}
    & & \textbf{0} & \textbf{1} & \textbf{3} & \textbf{5} \\  
    \midrule
    \rowcolor{black!10!} \multicolumn{6}{c}{\textbf{\textit{few-shot}}} \\
    \multirow{2}{*}{GPT-3.5} & STL & \colorbox{blue!10}{0.650} & 0.586\textbf{{\color{gray}\fontsize{5.5}{8.4}\selectfont(-0.064) }} & 0.593\textbf{{\color{gray}\fontsize{5.5}{8.4}\selectfont(-0.057) }} & 0.548\textbf{{\color{gray}\fontsize{5.5}{8.4}\selectfont(-0.102) }} \\  
                              & UTL & \colorbox{blue!10}{0.645} & 0.566\textbf{{\color{gray}\fontsize{5.5}{8.4}\selectfont(-0.079) }} & 0.564\textbf{{\color{gray}\fontsize{5.5}{8.4}\selectfont(-0.081) }} & 0.541\textbf{{\color{gray}\fontsize{5.5}{8.4}\selectfont(-0.104) }} \\  
    \cmidrule(lr){2-6}
    \multirow{2}{*}{GPT-4o} & STL & \colorbox{blue!10}{0.670} & 0.608\textbf{{\color{gray}\fontsize{5.5}{8.4}\selectfont(-0.062) }} & 0.602\textbf{{\color{gray}\fontsize{5.5}{8.4}\selectfont(-0.068) }} & 0.595\textbf{{\color{gray}\fontsize{5.5}{8.4}\selectfont(-0.075) }} \\  
                              & UTL & \colorbox{blue!10}{0.659} & 0.597\textbf{{\color{gray}\fontsize{5.5}{8.4}\selectfont(-0.062) }} & 0.607\textbf{{\color{gray}\fontsize{5.5}{8.4}\selectfont(-0.052) }} & 0.601\textbf{{\color{gray}\fontsize{5.5}{8.4}\selectfont(-0.058) }} \\  
    \cmidrule(lr){2-6} 
    \multirow{2}{*}{LLaMA3} & STL & \colorbox{blue!10}{0.640} & 0.599\textbf{{\color{gray}\fontsize{5.5}{8.4}\selectfont(-0.041) }} & 0.606\textbf{{\color{gray}\fontsize{5.5}{8.4}\selectfont(-0.034) }} & 0.603\textbf{{\color{gray}\fontsize{5.5}{8.4}\selectfont(-0.037) }} \\  
                                & UTL & \colorbox{blue!10}{0.640} & 0.582\textbf{{\color{gray}\fontsize{5.5}{8.4}\selectfont(-0.058) }} & 0.601\textbf{{\color{gray}\fontsize{5.5}{8.4}\selectfont(-0.039) }} & 0.597\textbf{{\color{gray}\fontsize{5.5}{8.4}\selectfont(-0.043) }} \\
    \midrule
  
    Avg\textbf{{\color{gray}\fontsize{5.5}{8.4}\selectfont(\textpm stdev.) }} &  & \colorbox{blue!10}{0.650\textbf{{\color{gray}\fontsize{5.5}{8.4}\selectfont(\textpm 0.011) }}} & 0.589\textbf{{\color{gray}\fontsize{5.5}{8.4}\selectfont(\textpm 0.014) }}& 0.595\textbf{{\color{gray}\fontsize{5.5}{8.4}\selectfont(\textpm 0.016) }}& 0.580\textbf{{\color{gray}\fontsize{5.5}{8.4}\selectfont(\textpm 0.028) }} \\

    \rowcolor{black!10!} \multicolumn{6}{c}{\textbf{\textit{CoT}}} \\
    \addlinespace[0.7mm]
    \multirow{2}{*}{GPT-3.5} & STL & 0.653\textbf{{\color{gray}\fontsize{5.5}{8.4}\selectfont(+0.003) }} & 0.602\textbf{{\color{gray}\fontsize{5.5}{8.4}\selectfont(-0.048) }} & 0.544\textbf{{\color{gray}\fontsize{5.5}{8.4}\selectfont(-0.106) }} & 0.533\textbf{{\color{gray}\fontsize{5.5}{8.4}\selectfont(-0.117) }} \\  
                              & UTL & 0.650\textbf{{\color{gray}\fontsize{5.5}{8.4}\selectfont(+0.005) }} & 0.575\textbf{{\color{gray}\fontsize{5.5}{8.4}\selectfont(-0.007) }} & 0.548\textbf{{\color{gray}\fontsize{5.5}{8.4}\selectfont(-0.097) }} & 0.516\textbf{{\color{gray}\fontsize{5.5}{8.4}\selectfont(-0.129) }} \\  
    \cmidrule(lr){2-6}
    \multirow{2}{*}{GPT-4o} & STL & 0.655\textbf{{\color{gray}\fontsize{5.5}{8.4}\selectfont(-0.015) }} & 0.615\textbf{{\color{gray}\fontsize{5.5}{8.4}\selectfont(-0.055) }} & 0.612\textbf{{\color{gray}\fontsize{5.5}{8.4}\selectfont(-0.058) }} & 0.605\textbf{{\color{gray}\fontsize{5.5}{8.4}\selectfont(-0.065) }} \\  
                              & UTL & 0.659\textbf{{\color{gray}\fontsize{5.5}{8.4}\selectfont(0) }} & 0.614\textbf{{\color{gray}\fontsize{5.5}{8.4}\selectfont(-0.045) }} & 0.611\textbf{{\color{gray}\fontsize{5.5}{8.4}\selectfont(-0.048) }} & 0.607\textbf{{\color{gray}\fontsize{5.5}{8.4}\selectfont(-0.052) }} \\  
    \cmidrule(lr){2-6} 
    \multirow{2}{*}{LLaMA3} & STL & 0.635\textbf{{\color{gray}\fontsize{5.5}{8.4}\selectfont(-0.005) }} & 0.603\textbf{{\color{gray}\fontsize{5.5}{8.4}\selectfont(-0.037) }} & 0.613\textbf{{\color{gray}\fontsize{5.5}{8.4}\selectfont(-0.027) }} & 0.610\textbf{{\color{gray}\fontsize{5.5}{8.4}\selectfont(-0.003) }} \\  
                                & UTL & 0.644\textbf{{\color{gray}\fontsize{5.5}{8.4}\selectfont(+0.004) }} & 0.586\textbf{{\color{gray}\fontsize{5.5}{8.4}\selectfont(-0.054) }} & 0.601\textbf{{\color{gray}\fontsize{5.5}{8.4}\selectfont(-0.039) }} & 0.598\textbf{{\color{gray}\fontsize{5.5}{8.4}\selectfont(-0.042) }} \\  
    \midrule

    Avg\textbf{{\color{gray}\fontsize{5.5}{8.4}\selectfont(\textpm stdev.) }} &  & 0.649\textbf{{\color{gray}\fontsize{5.5}{8.4}\selectfont(\textpm 0.008) }} & 0.599\textbf{{\color{gray}\fontsize{5.5}{8.4}\selectfont(\textpm 0.015) }}& 0.588\textbf{{\color{gray}\fontsize{5.5}{8.4}\selectfont(\textpm 0.032) }}& 0.578\textbf{{\color{gray}\fontsize{5.5}{8.4}\selectfont(\textpm 0.042) }}\\
        \bottomrule

    \end{tabular}
        \rmspace
\end{table}
Table~\ref{tab:prompt_type_with_examples} shows the impact of prompt type and number of examples on model performance for STL and UTL. Surprisingly, in-context demonstrations do not enhance performance for few-shot or CoT experiments (see Appendix~\ref{appendix:observations} for detailed insights). For both experiments, the setting with zero examples achieves the highest average performance: few-shot 0.650 (±0.011) and CoT 0.649 (±0.008). 
Performance consistently declines as the number of examples increase, likely due to noise that impairs generalization. \textit{Overall, there is no significant difference between few-shot and CoT.}

\subsection{Output Refinement}

We examine two output refinement strategies, Schema Mapping and Data Cleaning (see Sec.~\ref{subsec:post_processing}), to evaluate their impact shown in Table~\ref{tab:post_processing}.

\noindent \textbf{Schema Mapping} involves mapping the predicted schema keys to the target schema keys. Our results show no change in F1 scores compared to the initial predictions. This suggests that the models already effectively return the correct attributes, making the mapping step unnecessary.

\noindent \textbf{Data Cleaning} uses the defined data types to perform automatic value cleaning, consistently achieving the highest F1 scores across all models. \emph{This underscores the need for post-processing steps for IE with LLMs to align the extracted data with the target format to handle LLM hallucinations and inconsistent source data formats (see Sec.~\ref{subsec:post_processing}).}
\begin{table}[htbp!]
    \small
    \caption{Different LLMs across \textbf{STL} and \textbf{UTL} levels with different post-processing strategies. Baseline configuration is highlighted in \colorbox{blue!10}{light blue}.} 
    \label{tab:post_processing}
    \resizebox{\linewidth}{!}{
    \setlength{\tabcolsep}{4pt}
    \begin{tabular}{l c c c c}
    \toprule
    \multirow{2}{*}{\textbf{Models}} & \multirow{2}{*}{\textbf{Level}} & \multicolumn{3}{c}{\textbf{Exact Match (F1)}} \\
    \cmidrule(lr){3-5}
    & & \textbf{Initial Pred.} & \textbf{Mapped Pred.} & \textbf{Cleaned Pred.} \\
    \midrule
    \multirow{2}{*}{GPT-3.5} & STL & \colorbox{blue!10}{0.650} & 0.650\textbf{{\color{gray}\fontsize{5.5}{8.4}\selectfont(0) }} & 0.737\textbf{{\color{gray}\fontsize{5.5}{8.4}\selectfont(+0.087) }} \\
             & UTL & \colorbox{blue!10}{0.645} & 0.645\textbf{{\color{gray}\fontsize{5.5}{8.4}\selectfont(0) }} & 0.733\textbf{{\color{gray}\fontsize{5.5}{8.4}\selectfont(+0.088) }} \\
    \cmidrule(lr){2-5}
    \multirow{2}{*}{GPT-4o} & STL & \colorbox{blue!10}{0.670} & 0.670\textbf{{\color{gray}\fontsize{5.5}{8.4}\selectfont(0) }} & 0.749\textbf{{\color{gray}\fontsize{5.5}{8.4}\selectfont(+0.079) }} \\
             & UTL & \colorbox{blue!10}{0.659} & 0.659\textbf{{\color{gray}\fontsize{5.5}{8.4}\selectfont(0) }} & 0.741\textbf{{\color{gray}\fontsize{5.5}{8.4}\selectfont(+0.082) }} \\
    \cmidrule(lr){2-5}
    \multirow{2}{*}{LLaMA3} & STL & \colorbox{blue!10}{0.640} & 0.640\textbf{{\color{gray}\fontsize{5.5}{8.4}\selectfont(0) }} & 0.724\textbf{{\color{gray}\fontsize{5.5}{8.4}\selectfont(+0.084) }} \\
               & UTL & \colorbox{blue!10}{0.640} & 0.640\textbf{{\color{gray}\fontsize{5.5}{8.4}\selectfont(0) }} & 0.725\textbf{{\color{gray}\fontsize{5.5}{8.4}\selectfont(+0.085) }} \\
    \bottomrule
    Avg\textbf{{\color{gray}\fontsize{5.5}{8.4}\selectfont(\textpm stdev.) }} & & \colorbox{blue!10}{0.650\textbf{{\color{gray}\fontsize{5.5}{8.4}\selectfont(\textpm 0.011) }}} & 0.650\textbf{{\color{gray}\fontsize{5.5}{8.4}\selectfont(\textpm 0.011) }} & 0.734\textbf{{\color{gray}\fontsize{5.5}{8.4}\selectfont(\textpm 0.009) }}
    \end{tabular}}
        \rmspace
\end{table}
\subsection{Evaluation Techniques}

We explore three evaluation techniques to assess their impact on model performance (see Sec.~\ref{subsec:evaluation_techniques}).
On average, Fuzzy Match achieved the highest F1 score (0.733), outperforming Substring Match (0.676) and Exact Match, as shown in Table~\ref{tab:evaluation_techniques}. 

\noindent \final{To validate the reliability of these metrics, we manually analyzed cases where Exact Match failed but Substring or Fuzzy Match succeeded. This review confirmed that most soft-matched predictions were semantically correct, with errors largely due to OCR noise, formatting differences, or annotation inconsistencies. Notably, fuzzy match (with a strict 0.8 threshold) captured valid predictions that exact and substring match missed, such as truncated organization names. Our detailed error analysis in Appendix~\ref{appendix:eval_tech} shows that fuzzy and substring matching provide near-perfect precision when manually checked for semantic equivalence, with precision scores of 0.98 and 1.00, respectively.}
\textit{This shows Fuzzy Match's ability to balance flexibility and precision. 
}

\begin{table}[hbtp]
    \small
    \caption{ Different LLMs across \textbf{STL} and \textbf{UTL} levels with different evaluation techniques. Baseline configuration in \colorbox{blue!10}{light blue}.}
    \label{tab:evaluation_techniques}
    \resizebox{\linewidth}{!}{
    \setlength{\tabcolsep}{4pt}
    \begin{tabular}{l c c c c}
    \toprule
    \multirow{2}{*}{\textbf{Models}} & \multirow{2}{*}{\textbf{Level}} & \textbf{Exact Match} & \textbf{Substring Match} & \textbf{Fuzzy Match} \\
    & & \textbf{(F1)} & \textbf{(F1)} & \textbf{(F1)} \\
    \midrule
    \multirow{2}{*}{GPT-3.5} & STL & \colorbox{blue!10}{0.650} & 0.683\textbf{{\color{gray}\fontsize{5.5}{8.4}\selectfont(+0.033) }} & 0.730\textbf{{\color{gray}\fontsize{5.5}{8.4}\selectfont(+0.080) }} \\
             & UTL & \colorbox{blue!10}{0.645} & 0.682\textbf{{\color{gray}\fontsize{5.5}{8.4}\selectfont(+0.037) }} & 0.726\textbf{{\color{gray}\fontsize{5.5}{8.4}\selectfont(+0.081) }} \\
    \cmidrule(lr){2-5}
    \multirow{2}{*}{GPT-4o} & STL & \colorbox{blue!10}{0.670} & 0.690\textbf{{\color{gray}\fontsize{5.5}{8.4}\selectfont(+0.020) }} & 0.750\textbf{{\color{gray}\fontsize{5.5}{8.4}\selectfont(+0.080) }} \\
             & UTL & \colorbox{blue!10}{0.659} & 0.678\textbf{{\color{gray}\fontsize{5.5}{8.4}\selectfont(+0.019) }} & 0.744\textbf{{\color{gray}\fontsize{5.5}{8.4}\selectfont(+0.085) }} \\
    \cmidrule(lr){2-5}
    \multirow{2}{*}{LLaMA3} & STL & \colorbox{blue!10}{0.640} & 0.661\textbf{{\color{gray}\fontsize{5.5}{8.4}\selectfont(+0.021) }} & 0.727\textbf{{\color{gray}\fontsize{5.5}{8.4}\selectfont(+0.087) }} \\
               & UTL & \colorbox{blue!10}{0.640} & 0.662\textbf{{\color{gray}\fontsize{5.5}{8.4}\selectfont(+0.022) }} & 0.723\textbf{{\color{gray}\fontsize{5.5}{8.4}\selectfont(+0.083) }} \\
    \bottomrule
    Avg\textbf{{\color{gray}\fontsize{5.5}{8.4}\selectfont(\textpm stdev.) }} & & \colorbox{blue!10}{0.650\textbf{{\color{gray}\fontsize{5.5}{8.4}\selectfont(\textpm 0.011) }}} & 0.676\textbf{{\color{gray}\fontsize{5.5}{8.4}\selectfont(\textpm 0.011) }} & 0.733\textbf{{\color{gray}\fontsize{5.5}{8.4}\selectfont(\textpm 0.010) }}
    \end{tabular}}
    \rmspace
\end{table}

\subsection{Putting it All Together}

\begin{figure}[ht]
    \centering
    \includegraphics[width=\columnwidth]{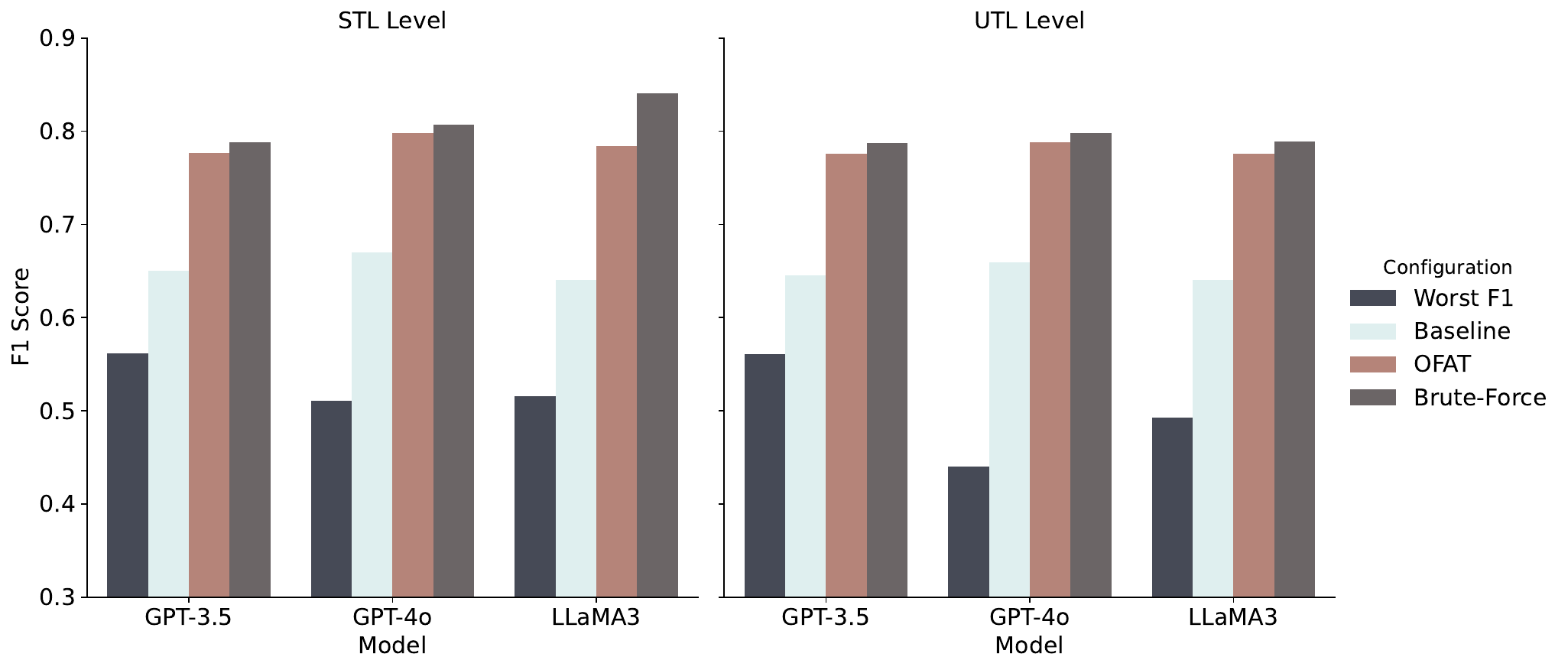}
    \caption{F1 scores of different LLMs across three configurations (Baseline, OFAT, and Brute-Force). Each bar represents the mean F1 score of a model for the corresponding configuration. \final{The OFAT configuration uses a single global parameter set optimized across all models.}}
    \label{fig:configuration_comparison}
\rmspace
\end{figure}

We investigated the influence of various parameters on model performance along the IE extraction pipeline, analyzing one factor at a time. Drawing from the underlying 12 experiments, we identified the optimal parameter for each step and each model based on the experimental outcomes (Table~\ref{tab:OFAT_configurations_per_model}). \final{To validate this approach}, we conducted an exhaustive full factorial exploration with 432 configurations to identify \final{the overall best-performing (Table~\ref{tab:Brute_force_configurations_per_model}) and worst-performing parameter combinations per LLM.}
\final{For the comparative analysis in Figure~\ref{fig:configuration_comparison}, we use a single global OFAT configuration that achieves the highest average performance across all models (selection criteria in Appendix~\ref{appendix:observations}).}
The performance of these different configurations, including the worst, is depicted in Figure~\ref{fig:configuration_comparison}. We gain several insights: 

\begin{itemize}[leftmargin=*,noitemsep,topsep=0pt]

\item \emph{OFAT approximates well the best-performing Brute-Force configuration with a fraction ($\sim2.8\%$) of the required computation.} We see in Table~\ref{tab:OFAT_configurations_per_model} and Table~\ref{tab:Brute_force_configurations_per_model} that they match except for 8 parameter choices (\texttt{chunk size}, \texttt{prompt} and \texttt{example No.} parameters). Likewise, their F1 scores are close to each other (Figure~\ref{fig:configuration_comparison}), with OFAT achieving 0.783 and Brute-Force achieving 0.801 overall.

\item \emph{Adapting the IE pipeline to the LLM is crucial for competitive performance.} 
Overall, the OFAT configuration improves from baseline F1 of 0.650 to 0.783, a $20\%$ improvement. For the best-performing Brute-Force, the improvement is $23\%$. In comparison, the worst configuration achieves an average score of only 0.5, which is approximately 64.1\% of the best configuration’s performance.

\item \emph{Common patterns across LLMs.} First, output refinement and evaluation techniques boost performance for all LLMs. Second, there is a trend towards larger context sizes. Lastly, larger models (GPT-4o, LLaMa3) benefit more from examples, while the CoT pattern generally aids IE.

\end{itemize}

\added{\noindent \textbf{FUNSD Dataset.} We provide the equivalent OFAT analysis for the FUNSD dataset in Appendix~\ref{appendix:funsd_results}. 
While we observed different patterns with respect to the IE pipeline configuration, our experiments on FUNSD also show a clear improvement through 
\final{output refinement techniques and the use of tolerant evaluation metrics.}
Most importantly, they show that our OFAT method nearly reaches the best-performing Brute-Force performance (only $0.8$ F1 points lower), and improves 90\% from the baseline IE results. \emph{This indicates that our OFAT method for IE pipeline configuration is generalizable and can help practitioners and researchers to find the best configuration for their specific dataset.}}

\begin{table}[hbtp]
\centering
\footnotesize
\caption{OFAT configurations on a per-model basis and corresponding performance results.}
\begin{tabular}{lcccc}
\toprule
\textbf{Parameter} & \textbf{GPT-3.5} & \textbf{GPT-4o} & \textbf{LLaMA3} \\
\midrule
Input Type & Markdown & OCR & Markdown  \\
Chunk Size & Medium & Max & Max \\
Prompt & CoT & Few-Shot & Few-Shot &\\
Example No. & 0 & 0 & 0 \\
Output Refin. & Cleaned & Cleaned & Cleaned \\
Evaluation & Fuzzy & Fuzzy & Fuzzy \\
\bottomrule
\end{tabular}
\label{tab:OFAT_configurations_per_model}
\rmspace
\end{table}

\begin{table}[hbtp]
\centering
\footnotesize
\caption{\final{Best-performing} Brute-Force configurations on a per-model basis and corresponding performance results.}
\begin{tabular}{lcccc}
\toprule
\textbf{Parameter} & \textbf{GPT-3.5} & \textbf{GPT-4o} & \textbf{LLaMA3} \\
\midrule
Input Type & Markdown & OCR & Markdown  \\
Chunk Size & Max & Medium & Small \\
Prompt & Few-shot & CoT & CoT &\\
Example No. & 0 & 5 & 5 \\
Output Refin. & Cleaned & Cleaned & Cleaned \\
Evaluation & Fuzzy & Fuzzy & Fuzzy \\
\bottomrule
\end{tabular}
\label{tab:Brute_force_configurations_per_model}
\rmspace
\end{table}

Finally, we compare our text-based approach to IE to (1) GPT-4o-vision \final{and Qwen2.5-vision, multimodal LLMs}, (2) \added{LayoutLMv3 and ERNIE-Layout}, layout-aware models that we fine-tune to our dataset (Table~\ref{tab:model_comparison}).
Despite fine-tuning, \added{LayoutLMv3 and ERNIE-Layout} perform worse than our evaluated LLMs, even on documents that contain the same structure as the fine-tuning data (STL). We also observe that LLMs remain stable across template types. The best-performing \final{models are GPT-4o-vision followed by Qwen2.5-vision, both multimodal LLMs}, where we directly provide an image of the PDF for IE. These models benefits from less context loss due to chunking and their ability to use textual and visual features jointly. However, when considering token usage and cost, GPT-4o-vision requires $\sim2\times$ the tokens compared to text-only approaches and $>10\times$ the cost under the current OpenAI pricing scheme (Nov. 2024), \final{while Qwen2.5-vision uses $\sim1.4\times$ the tokens with no API costs when run locally.} Therefore, text-only approaches constitute a good trade-off between performance and cost for practical applications, \final{with Qwen2.5-vision offering a viable middle ground}.

\begin{table}[hbtp]
    \centering %
    \footnotesize %
    \caption{Cost (per experiment) and performance of LLMs across \textbf{STL} and \textbf{UTL} levels for the OFAT configuration. The tokens for LayoutLMv3 and ERNIE-Layout include the fine-tuning. $^*$We run Qwen2.5-vision, LayoutLMv3, \added{ERNIE-Layout}, and LlaMa3 locally, without API costs. } 
    \label{tab:model_comparison}
    \setlength{\tabcolsep}{3pt}
    \begin{tabular}{l r l l l}
    \toprule
    \textbf{Models} & \textbf{Tokens} & \textbf{API Cost} & \textbf{STL (F1)} & \textbf{UTL (F1)} \\
    \midrule
    GPT-3.5                          & 322K  & \$0.18  & 0.777  & 0.776  \\ 
    GPT-4o                           & 278K  & \$0.40  & 0.798  & 0.788  \\
    LLaMA3                           & 239K  & \$0$^*$      & 0.784  & 0.776  \\
    GPT-4o-vision                     & 585K  & \$4.76   & 0.902  & 0.897  \\
    \added{Qwen2.5-vision}                     & \added{385K}  & \added{\$0$^*$}   & \added{0.890} & \added{0.891}  \\
    LayoutLMv3                       & 176.1M  & \$0$^*$      & 0.603  & 0.194  \\
    ERNIE-Layout & 240M  & \$0$^*$ & 0.663  & 0.397  \\
    \bottomrule
    \end{tabular}
    \rmspace
\end{table}

\section{Related Work}
\label{sec:related_work}

\noindent \textbf{IE using LLMs.} Transformer-based models~\cite{devlin2019bert, brown2020language, liu2019roberta} have advanced NLP with self-attention and large-scale pretraining but struggle with layout-rich documents (LRDs). To address this, layout-aware and multimodal models have emerged~\cite{xu2020layoutlm, xu2020layoutlmv2, huang2022layoutlmv3}, \final{including generative approaches~\cite{powalski2021going}}. Additionally, some models enhance document understanding through multimodal learning~\cite{openai2023gpt4, anil2023gemini}. Structural-aware approaches~\cite{fujitake-2024-layoutllm, lee2022formnetstructuralencodingsequential} further improve extraction accuracy, while end-to-end models~\cite{kim2022ocr} bypass OCR for direct document image processing. Furthermore, some approaches adopt a chat-based paradigm for flexible IE~\cite{xu-etal-2024-chatuie}, while others~\cite{jiang-etal-2025-relayout} enhance document understanding through layout-aware pretraining, advancing LLMs for practical applications. 

\noindent \textbf{Strategies for IE from LRDs.} Graph-based models~\cite{liu-etal-2019-graph, nourbakhsh-etal-2024-aligatr} enhance relation extraction by capturing textual-visual relationships. A critical factor in layout-rich document (LRD) processing is maintaining the reading order, where Token Path Prediction (TPP)\cite{zhang-etal-2023-reading} resolves OCR layout ambiguities, and global tagging\cite{shaojie-etal-2023-document} mitigates text ordering issues, thereby improving extraction accuracy. For structured data, specific approaches target various document types: table extraction~\cite{shigarov2018tabbypdf} and unified OCR, preprocessing, and postprocessing for document IE~\cite{perot-etal-2024-lmdx}. Additionally, a simple yet effective multimodal and multilingual method for parsing semi-structured forms is presented in~\cite{cheng-etal-2025-xformparser}. \added{Despite the variety of models proposed, comparative studies evaluating text-only, multimodal, and layout-aware LLMs remain limited~\cite{van2023document, stanislawek2021kleister}}.

\noindent \textbf{Preprocessing, Chunking, Prompting, Postprocessing, and Evaluation Techniques.} Chain-of-Thought (CoT) prompting~\cite{wei2022cot} enhances reasoning in complex LRD extraction tasks, while diverse prompt-response datasets~\cite{zmigrod-etal-2024-value} improve LLM robustness. Instruction-finetuned LLMs have also demonstrated effectiveness in domain-specific applications, such as clinical and biomedical tasks, where zero-shot and few-shot enable adaptive extraction without extensive fine-tuning~\cite{labrak-etal-2024-zero}. Postprocessing techniques, including text normalization and schema alignment~\cite{furst2023versamatch}.

Despite advances, current studies often focus on isolated components rather than evaluating complete IE pipelines, leading to an incomplete understanding of their interplay. \added{Our findings, quantified and contextualized within a controlled and reproducible framework. Our study enables structured evaluation of IE techniques and systematically analyzes the trade-offs between performance and efficiency across diverse model categories.}

\section{Conclusion}
\label{sec:conclusion}

This paper introduces the design space for IE from LRDs using LLMs, with challenges of data structuring, model engagement, and output refinement, considering the design choices within those challenges. We examine the effects of these choices on the overall performance through one-factor-at-a-time (OFAT) and full-factorial (Brute-Force) exploration of the parameters. We show that well-configured general-purpose LLMs provide an effective alternative to specialized models for IE from LRDs.
\added{Our cost-effective OFAT method (2.8\% of required computation) generalizes across two datasets (VRDU and FUNSD) with a performance that is just 0.8--1.8\% lower than the best-possible configuration that can be found through a full-factorial exploration. This makes our results and method directly applicable to IE practitioners who need to configure their IE pipeline for their dataset.} Towards this goal, we have open-sourced LayIE-LLM, our layout-aware IE test suite (\url{https://github.com/gayecolakoglu/LayIE-LLM}).

\section*{Acknowledgments}
This work was partially supported by DataGEMS, funded by the European Union’s Horizon Europe Research and Innovation Programme, under grant agreement No 101188416.

\clearpage
\section*{Limitations}
\label{sec:limitations}

Our experimental evaluation focused on the Visually Rich Document Understanding (VRDU) dataset \cite{Wang_2023} \added{and the FUNSD dataset \cite{jaume2019funsd}. VRDU was chosen for its comprehensive OCR outputs, well-defined schema, and diverse template types (Amendment, Short-Form, and Dissemination) that capture complex layouts. 
It also supports generalization tasks--Single Template Learning (STL), Unseen Template Learning (UTL), and Mixed Template Learning (MTL)--making it suitable for visually rich document understanding.
In contrast, FUNSD was selected due to its widespread use in document understanding tasks, particularly for extracting structured information from simpler, form-based layouts. While FUNSD does not share the layout diversity and complexity of VRDU, it serves as a useful baseline for form-centric document analysis. This provides a different perspective by offering insights into scenarios where key-value pairs are organized within relatively simple templates.}

\added{One key limitation arises from the differences in schema structure between these datasets. The VRDU dataset has structured templates that facilitate consistent key mapping during post-processing (see Section~\ref{subsec:post_processing}), while the FUNSD dataset features dynamic, document-specific key-value pairs, making mapping more challenging. To address this, we plan to enhance our existing schema-mapping step by incorporating context-aware strategies that dynamically adapt to variations and leverage semantic similarity to reduce mismatches.}

\added{Another limitation lies in the construction of few-shot prompts (see Appendix~\ref{appendix:dataset_details}). Currently, the LLM generates these prompts by condensing the original OCR data while preserving only text features, aiming to reduce token usage and cost. However, in the tested document, both text and layout features are preserved without modification, embedding spatial structure into text sequences to enhance model understanding. Improving the LLM-generated few-shot prompt process by better integrating spatial features directly into the prompts could enhance the model's performance in complex scenarios.}

\added{Our manual investigation (as detailed in Appendix~\ref{appendix:observations}) demonstrated that fuzzy match is more tolerant of valid variations while maintaining high accuracy, as verified through manual evaluation; however, we acknowledge that it may relax correctness criteria and inflate performance metrics. To address this,} we plan to extend the evaluation with metrics commonly used in the context of LLMs, such as ROUGE~\cite{lin2004rouge} and BLEU~\cite{papineni2002bleu} scores and their adaptations~\cite{yang2018adaptations}. A fair comparison using cost-related metrics is mostly harder to compute due to their changing depending on the scenario. For instance, for Llama3, ERNIE-Layout, and LayoutLMv3, we listed the cost as 0 due to having a free-to-use service; it comes with a cost. Similarly, the cost of computation can be considered from different perspectives, such as the energy usage of a model. 

The test suite is currently limited to the steps that we listed in Fig.~\ref{fig:pipeline}, whereas one could imagine additional steps or factors that affect the performance and may even deliver more satisfactory outcomes. We would like to incorporate additional steps that we learn from the community and incrementally enlarge the design space, and extend the testing capabilities for IE from LRDs.

\added{Another limitation stems from the OFAT strategy used to explore the parameter space. While OFAT enables systematic tuning with minimal computational cost, it inherently fails to detect interaction effects between parameter combinations. To evaluate whether this limitation affected our results, we performed a brute-force search as a benchmark, directly comparing the OFAT outcomes to the best results from the brute-force approach. The comparison showed that OFAT approximated the full-grid results well, indicating that, in our specific scenario, OFAT was sufficient despite its theoretical limitations. We consider this trade-off acceptable for our goal of providing a fast, reproducible tuning method for real-world LLM deployment. Nevertheless, future work may incorporate interaction-aware designs to further uncover parameter synergies.}

Last limitation of our study is the variability in performance between the different types of models we tested when handling LRDs. \added{While some models, such as LayoutLMv3 and ERNIE-Layout}, are fine-tuned for layout-aware tasks, others, like LLaMA3 and GPT-4o-vision, are optimized for text processing or multi-modal inputs, with their relative strengths and weaknesses varying across the evaluated dimensions. This underscores the need for broader benchmarking with additional models, such as DocLLM, LayoutLLM, and ReLayout, to better understand performance differences.

\section*{Ethical Considerations}
Large Language Models (LLMs) can contain biases that can have a negative impact on marginalized groups~\cite{gallegos2024bias}. For the task of information extraction, this could have the impact that uncommon names for people and places are auto-corrected by the LLM to their more common form. In our experiments, we have encountered some instances of this and plan to investigate this further.

\bibliography{bibliography.bib}

\begin{thebibliography}{60}
\providecommand{\natexlab}[1]{#1}

\bibitem[{Appalaraju et~al.(2021)Appalaraju, Jasani, Kota, Xie, and Manmatha}]{appalaraju2021docformer}
Srikar Appalaraju, Bhavan Jasani, Bhargava~Urala Kota, Yusheng Xie, and R~Manmatha. 2021.
\newblock Docformer: End-to-end transformer for document understanding.
\newblock In \emph{Proceedings of the IEEE/CVF international conference on computer vision}, pages 993--1003.

\bibitem[{Bai et~al.(2024)Bai, Kang, Stanovsky, Freitag, Dredze, and Ritter}]{bai-etal-2024-schema}
Fan Bai, Junmo Kang, Gabriel Stanovsky, Dayne Freitag, Mark Dredze, and Alan Ritter. 2024.
\newblock \href {https://doi.org/10.18653/v1/2024.findings-emnlp.600} {Schema-driven information extraction from heterogeneous tables}.
\newblock In \emph{Findings of the Association for Computational Linguistics: EMNLP 2024}, pages 10252--10273, Miami, Florida, USA. Association for Computational Linguistics.

\bibitem[{Bai et~al.(2025)Bai, Chen, Liu, Wang, Ge, Song, Dang, Wang, Wang, Tang et~al.}]{bai2025qwen2}
Shuai Bai, Keqin Chen, Xuejing Liu, Jialin Wang, Wenbin Ge, Sibo Song, Kai Dang, Peng Wang, Shijie Wang, Jun Tang, et~al. 2025.
\newblock Qwen2.5-vl technical report.
\newblock \emph{arXiv preprint arXiv:2502.13923}.

\bibitem[{Bhadauria et~al.(2024)Bhadauria, Sierra~M{\'u}nera, and Krestel}]{bhadauria-etal-2024-effects}
Divya Bhadauria, Alejandro Sierra~M{\'u}nera, and Ralf Krestel. 2024.
\newblock \href {https://aclanthology.org/2024.wnut-1.8/} {The effects of data quality on named entity recognition}.
\newblock In \emph{Proceedings of the Ninth Workshop on Noisy and User-generated Text (W-NUT 2024)}, pages 79--88, San {\.{G}}iljan, Malta. Association for Computational Linguistics.

\bibitem[{Biten et~al.(2019)Biten, Tito, Mafla, Gomez, Rusinol, Valveny, Jawahar, and Karatzas}]{biten2019scene}
Ali~Furkan Biten, Ruben Tito, Andres Mafla, Lluis Gomez, Mar{\c{c}}al Rusinol, Ernest Valveny, CV~Jawahar, and Dimosthenis Karatzas. 2019.
\newblock Scene text visual question answering.
\newblock In \emph{Proceedings of the IEEE/CVF international conference on computer vision}, pages 4291--4301.

\bibitem[{Blau et~al.(2024)Blau, Kimhi, Belinkov, Bronstein, and Baskin}]{blau2024context}
Tsachi Blau, Moshe Kimhi, Yonatan Belinkov, Alexander Bronstein, and Chaim Baskin. 2024.
\newblock Context-aware prompt tuning: Advancing in-context learning with adversarial methods.
\newblock \emph{arXiv preprint arXiv:2410.17222}.

\bibitem[{Brown et~al.(2020)Brown, Mann, Ryder, Subbiah, Kaplan, Dhariwal, Neelakantan, Shyam, Sastry, Askell et~al.}]{brown2020language}
Tom Brown, Benjamin Mann, Nick Ryder, Melanie Subbiah, Jared~D Kaplan, Prafulla Dhariwal, Arvind Neelakantan, Pranav Shyam, Girish Sastry, Amanda Askell, et~al. 2020.
\newblock Language models are few-shot learners.
\newblock \emph{Advances in neural information processing systems}, 33:1877--1901.

\bibitem[{Cheng et~al.(2025)Cheng, Zhang, Yang, Li, Zhou, Liu, Wu, Guan, Sun, Wu, Li, and Li}]{cheng-etal-2025-xformparser}
Xianfu Cheng, Hang Zhang, Jian Yang, Xiang Li, Weixiao Zhou, Fei Liu, Kui Wu, Xiangyuan Guan, Tao Sun, Xianjie Wu, Tongliang Li, and Zhoujun Li. 2025.
\newblock \href {https://aclanthology.org/2025.coling-main.41/} {{XF}orm{P}arser: A simple and effective multimodal multilingual semi-structured form parser}.
\newblock In \emph{Proceedings of the 31st International Conference on Computational Linguistics}, pages 606--620, Abu Dhabi, UAE. Association for Computational Linguistics.

\bibitem[{Cui et~al.(2021)Cui, Xu, Lv, and Wei}]{cui2021document}
Lei Cui, Yiheng Xu, Tengchao Lv, and Furu Wei. 2021.
\newblock Document ai: Benchmarks, models and applications.
\newblock \emph{arXiv preprint arXiv:2111.08609}.

\bibitem[{Devlin et~al.(2019)Devlin, Chang, Lee, and Toutanova}]{devlin2019bert}
Jacob Devlin, Ming-Wei Chang, Kenton Lee, and Kristina Toutanova. 2019.
\newblock Bert: Pre-training of deep bidirectional transformers for language understanding.
\newblock \emph{arXiv preprint arXiv:1810.04805}.

\bibitem[{Fujitake(2024)}]{fujitake-2024-layoutllm}
Masato Fujitake. 2024.
\newblock \href {https://aclanthology.org/2024.lrec-main.892/} {{L}ayout{LLM}: Large language model instruction tuning for visually rich document understanding}.
\newblock In \emph{Proceedings of the 2024 Joint International Conference on Computational Linguistics, Language Resources and Evaluation (LREC-COLING 2024)}, pages 10219--10224, Torino, Italia. ELRA and ICCL.

\bibitem[{F{\"u}rst et~al.(2023)F{\"u}rst, Fadel~Argerich, and Cheng}]{furst2023versamatch}
Jonathan F{\"u}rst, Mauricio Fadel~Argerich, and Bin Cheng. 2023.
\newblock Versamatch: ontology matching with weak supervision.
\newblock In \emph{49th Conference on Very Large Data Bases (VLDB), Vancouver, Canada, 28 August-1 September 2023}, volume~16, pages 1305--1318. Association for Computing Machinery.

\bibitem[{Gallegos et~al.(2024)Gallegos, Rossi, Barrow, Tanjim, Kim, Dernoncourt, Yu, Zhang, and Ahmed}]{gallegos2024bias}
Isabel~O Gallegos, Ryan~A Rossi, Joe Barrow, Md~Mehrab Tanjim, Sungchul Kim, Franck Dernoncourt, Tong Yu, Ruiyi Zhang, and Nesreen~K Ahmed. 2024.
\newblock Bias and fairness in large language models: A survey.
\newblock \emph{Computational Linguistics}, pages 1--79.

\bibitem[{Huang et~al.(2025)Huang, Yu, Ma, Zhong, Feng, Wang, Chen, Peng, Feng, Qin et~al.}]{huang2023survey}
Lei Huang, Weijiang Yu, Weitao Ma, Weihong Zhong, Zhangyin Feng, Haotian Wang, Qianglong Chen, Weihua Peng, Xiaocheng Feng, Bing Qin, et~al. 2025.
\newblock A survey on hallucination in large language models: Principles, taxonomy, challenges, and open questions.
\newblock \emph{ACM Transactions on Information Systems}, 43(2):1--55.

\bibitem[{Huang et~al.(2022)Huang, Lv, Cui, Lu, and Wei}]{huang2022layoutlmv3}
Yupan Huang, Tengchao Lv, Lei Cui, Yutong Lu, and Furu Wei. 2022.
\newblock Layoutlmv3: Pre-training for document ai with unified text and image masking.
\newblock In \emph{Proceedings of the 30th ACM International Conference on Multimedia}, pages 4083--4091.

\bibitem[{Jaume et~al.(2019)Jaume, Ekenel, and Thiran}]{jaume2019funsd}
Guillaume Jaume, Hazim~Kemal Ekenel, and Jean-Philippe Thiran. 2019.
\newblock Funsd: A dataset for form understanding in noisy scanned documents.
\newblock In \emph{2019 International Conference on Document Analysis and Recognition Workshops (ICDARW)}, volume~2, pages 1--6. IEEE.

\bibitem[{Ji et~al.(2023)Ji, Lee, Frieske, Yu, Su, Xu, Ishii, Bang, Madotto, and Fung}]{ji2023survey}
Ziwei Ji, Nayeon Lee, Rita Frieske, Tiezheng Yu, Dan Su, Yan Xu, Etsuko Ishii, Ye~Jin Bang, Andrea Madotto, and Pascale Fung. 2023.
\newblock Survey of hallucination in natural language generation.
\newblock \emph{ACM Computing Surveys}, 55(12):1--38.

\bibitem[{Jiang et~al.(2025)Jiang, Wang, Chen, and Nakashima}]{jiang-etal-2025-relayout}
Zhouqiang Jiang, Bowen Wang, Junhao Chen, and Yuta Nakashima. 2025.
\newblock \href {https://aclanthology.org/2025.coling-main.255/} {{R}e{L}ayout: Towards real-world document understanding via layout-enhanced pre-training}.
\newblock In \emph{Proceedings of the 31st International Conference on Computational Linguistics}, pages 3778--3793, Abu Dhabi, UAE. Association for Computational Linguistics.

\bibitem[{Kim et~al.(2022)Kim, Hong, Yim, Nam, Park, Yim, Hwang, Yun, Han, and Park}]{kim2022ocr}
Geewook Kim, Teakgyu Hong, Moonbin Yim, JeongYeon Nam, Jinyoung Park, Jinyeong Yim, Wonseok Hwang, Sangdoo Yun, Dongyoon Han, and Seunghyun Park. 2022.
\newblock Ocr-free document understanding transformer.
\newblock In \emph{European Conference on Computer Vision}, pages 498--517. Springer.

\bibitem[{Labrak et~al.(2024)Labrak, Rouvier, and Dufour}]{labrak-etal-2024-zero}
Yanis Labrak, Mickael Rouvier, and Richard Dufour. 2024.
\newblock \href {https://aclanthology.org/2024.lrec-main.185/} {A zero-shot and few-shot study of instruction-finetuned large language models applied to clinical and biomedical tasks}.
\newblock In \emph{Proceedings of the 2024 Joint International Conference on Computational Linguistics, Language Resources and Evaluation (LREC-COLING 2024)}, pages 2049--2066, Torino, Italia. ELRA and ICCL.

\bibitem[{Lee et~al.(2022)Lee, Li, Dozat, Perot, Su, Hua, Ainslie, Wang, Fujii, and Pfister}]{lee2022formnetstructuralencodingsequential}
Chen-Yu Lee, Chun-Liang Li, Timothy Dozat, Vincent Perot, Guolong Su, Nan Hua, Joshua Ainslie, Renshen Wang, Yasuhisa Fujii, and Tomas Pfister. 2022.
\newblock \href {https://arxiv.org/abs/2203.08411} {Formnet: Structural encoding beyond sequential modeling in form document information extraction}.
\newblock \emph{Preprint}, arXiv:2203.08411.

\bibitem[{Lin(2004)}]{lin2004rouge}
Chin-Yew Lin. 2004.
\newblock Rouge: A package for automatic evaluation of summaries.
\newblock In \emph{Text summarization branches out}, pages 74--81.

\bibitem[{Liu et~al.(2024)Liu, Lin, Hewitt, Paranjape, Bevilacqua, Petroni, and Liang}]{liu2024lost}
Nelson~F Liu, Kevin Lin, John Hewitt, Ashwin Paranjape, Michele Bevilacqua, Fabio Petroni, and Percy Liang. 2024.
\newblock Lost in the middle: How language models use long contexts.
\newblock \emph{Transactions of the Association for Computational Linguistics}, 12:157--173.

\bibitem[{Liu et~al.(2019{\natexlab{a}})Liu, Gao, Zhang, and Zhao}]{liu-etal-2019-graph}
Xiaojing Liu, Feiyu Gao, Qiong Zhang, and Huasha Zhao. 2019{\natexlab{a}}.
\newblock \href {https://doi.org/10.18653/v1/N19-2005} {Graph convolution for multimodal information extraction from visually rich documents}.
\newblock In \emph{Proceedings of the 2019 Conference of the North {A}merican Chapter of the Association for Computational Linguistics: Human Language Technologies, Volume 2 (Industry Papers)}, pages 32--39, Minneapolis, Minnesota. Association for Computational Linguistics.

\bibitem[{Liu et~al.(2019{\natexlab{b}})Liu, Ott, Goyal, Du, Joshi, Chen, Levy, Lewis, Zettlemoyer, and Stoyanov}]{liu2019roberta}
Yinhan Liu, Myle Ott, Naman Goyal, Jingfei Du, Mandar Joshi, Danqi Chen, Omer Levy, Mike Lewis, Luke Zettlemoyer, and Veselin Stoyanov. 2019{\natexlab{b}}.
\newblock Roberta: A robustly optimized bert pretraining approach.
\newblock \emph{arXiv preprint arXiv:1907.11692}.

\bibitem[{Luo et~al.(2024{\natexlab{a}})Luo, Shen, Zhu, Zheng, Yu, and Yao}]{luo2024layoutllm}
Chuwei Luo, Yufan Shen, Zhaoqing Zhu, Qi~Zheng, Zhi Yu, and Cong Yao. 2024{\natexlab{a}}.
\newblock Layoutllm: Layout instruction tuning with large language models for document understanding.
\newblock In \emph{Proceedings of the IEEE/CVF conference on computer vision and pattern recognition}, pages 15630--15640.

\bibitem[{Luo et~al.(2024{\natexlab{b}})Luo, Chen, Guo, Li, Zeng, Cai, and Li}]{luo2024asgeaexploitinglogicrules}
Yangyifei Luo, Zhuo Chen, Lingbing Guo, Qian Li, Wenxuan Zeng, Zhixin Cai, and Jianxin Li. 2024{\natexlab{b}}.
\newblock \href {https://arxiv.org/abs/2402.11000} {Asgea: Exploiting logic rules from align-subgraphs for entity alignment}.
\newblock \emph{Preprint}, arXiv:2402.11000.

\bibitem[{Mieskes and Schmunk(2019)}]{mieskes-schmunk-2019-ocr}
Margot Mieskes and Stefan Schmunk. 2019.
\newblock \href {https://aclanthology.org/W19-3633/} {{OCR} quality and {NLP} preprocessing}.
\newblock In \emph{Proceedings of the 2019 Workshop on Widening NLP}, pages 102--105, Florence, Italy. Association for Computational Linguistics.

\bibitem[{Nourbakhsh et~al.(2024)Nourbakhsh, Jin, Parekh, Shah, and Rose}]{nourbakhsh-etal-2024-aligatr}
Armineh Nourbakhsh, Zhao Jin, Siddharth Parekh, Sameena Shah, and Carolyn Rose. 2024.
\newblock \href {https://doi.org/10.18653/v1/2024.findings-emnlp.778} {{A}li{GAT}r: Graph-based layout generation for form understanding}.
\newblock In \emph{Findings of the Association for Computational Linguistics: EMNLP 2024}, pages 13309--13328, Miami, Florida, USA. Association for Computational Linguistics.

\bibitem[{OpenAI(2023)}]{openai2023gpt4}
OpenAI. 2023.
\newblock \href {https://www.openai.com/research/gpt-4} {Gpt-4 technical report}.
\newblock \emph{OpenAI}.

\bibitem[{{OpenAI}(2024)}]{openai2024gpt4o}
{OpenAI}. 2024.
\newblock \href {https://openai.com/index/hello-gpt-4o/} {Hello gpt-4o}.

\bibitem[{Papineni et~al.(2002)Papineni, Roukos, Ward, and Zhu}]{papineni2002bleu}
Kishore Papineni, Salim Roukos, Todd Ward, and Wei-Jing Zhu. 2002.
\newblock Bleu: a method for automatic evaluation of machine translation.
\newblock In \emph{Proceedings of the 40th annual meeting of the Association for Computational Linguistics}, pages 311--318.

\bibitem[{Park et~al.(2019)Park, Shin, Lee, Lee, Surh, Seo, and Lee}]{park2019cord}
Seunghyun Park, Seung Shin, Bado Lee, Junyeop Lee, Jaeheung Surh, Minjoon Seo, and Hwalsuk Lee. 2019.
\newblock Cord: a consolidated receipt dataset for post-ocr parsing.
\newblock In \emph{Workshop on Document Intelligence at NeurIPS 2019}.

\bibitem[{Peng et~al.(2022)Peng, Pan, Wang, Luo, Zhang, Huang, Hu, Yin, Chen, Zhang et~al.}]{peng2022ernie}
Qiming Peng, Yinxu Pan, Wenjin Wang, Bin Luo, Zhenyu Zhang, Zhengjie Huang, Teng Hu, Weichong Yin, Yongfeng Chen, Yin Zhang, et~al. 2022.
\newblock Ernie-layout: Layout knowledge enhanced pre-training for visually-rich document understanding.
\newblock \emph{arXiv preprint arXiv:2210.06155}.

\bibitem[{Perot et~al.(2024)Perot, Kang, Luisier, Su, Sun, Boppana, Wang, Wang, Mu, Zhang, Lee, and Hua}]{perot-etal-2024-lmdx}
Vincent Perot, Kai Kang, Florian Luisier, Guolong Su, Xiaoyu Sun, Ramya~Sree Boppana, Zilong Wang, Zifeng Wang, Jiaqi Mu, Hao Zhang, Chen-Yu Lee, and Nan Hua. 2024.
\newblock \href {https://doi.org/10.18653/v1/2024.findings-acl.899} {{LMDX}: Language model-based document information extraction and localization}.
\newblock In \emph{Findings of the Association for Computational Linguistics: ACL 2024}, pages 15140--15168, Bangkok, Thailand. Association for Computational Linguistics.

\bibitem[{Powalski et~al.(2021)Powalski, Borchmann, Jurkiewicz, Dwojak, Pietruszka, and Pa{\l}ka}]{powalski2021going}
Rafa{\l} Powalski, {\L}ukasz Borchmann, Dawid Jurkiewicz, Tomasz Dwojak, Micha{\l} Pietruszka, and Gabriela Pa{\l}ka. 2021.
\newblock Going full-tilt boogie on document understanding with text-image-layout transformer.
\newblock In \emph{Document Analysis and Recognition--ICDAR 2021: 16th International Conference, Lausanne, Switzerland, September 5--10, 2021, Proceedings, Part II 16}, pages 732--747. Springer.

\bibitem[{Radford et~al.(2021)Radford, Kim, Hallacy, Ramesh, Goh, Agarwal, Sastry, Askell, Mishkin, Clark et~al.}]{radford2021learning}
Alec Radford, Jong~Wook Kim, Chris Hallacy, Aditya Ramesh, Gabriel Goh, Sandhini Agarwal, Girish Sastry, Amanda Askell, Pamela Mishkin, Jack Clark, et~al. 2021.
\newblock Learning transferable visual models from natural language supervision.
\newblock In \emph{International conference on machine learning}, pages 8748--8763. PMLR.

\bibitem[{Shaojie et~al.(2023)Shaojie, Tianshu, Yaojie, Hongyu, Xianpei, Yingfei, and Le}]{shaojie-etal-2023-document}
He~Shaojie, Wang Tianshu, Lu~Yaojie, Lin Hongyu, Han Xianpei, Sun Yingfei, and Sun Le. 2023.
\newblock \href {https://aclanthology.org/2023.ccl-1.62/} {Document information extraction via global tagging}.
\newblock In \emph{Proceedings of the 22nd Chinese National Conference on Computational Linguistics}, pages 726--735, Harbin, China. Chinese Information Processing Society of China.

\bibitem[{Shigarov et~al.(2018)Shigarov, Altaev, Mikhailov, Paramonov, and Cherkashin}]{shigarov2018tabbypdf}
Alexey Shigarov, Andrey Altaev, Andrey Mikhailov, Viacheslav Paramonov, and Evgeniy Cherkashin. 2018.
\newblock Tabbypdf: Web-based system for pdf table extraction.
\newblock In \emph{International Conference on Information and Software Technologies}, pages 257--269. Springer.

\bibitem[{Smith(2007)}]{smith2007overview}
Ray Smith. 2007.
\newblock An overview of the tesseract ocr engine.
\newblock In \emph{Ninth international conference on document analysis and recognition (ICDAR 2007)}, volume~2, pages 629--633. IEEE.

\bibitem[{Srivastava and Dong(2013)}]{dong2013big}
Divesh Srivastava and Xin~Luna Dong. 2013.
\newblock Big data integration.
\newblock In \emph{2013 IEEE 29th international conference on data engineering (ICDE)}, pages 1245--1248. IEEE.

\bibitem[{Stanis{\l}awek et~al.(2021)Stanis{\l}awek, Grali{\'n}ski, Wr{\'o}blewska, Lipi{\'n}ski, Kaliska, Rosalska, Topolski, and Biecek}]{stanislawek2021kleister}
Tomasz Stanis{\l}awek, Filip Grali{\'n}ski, Anna Wr{\'o}blewska, Dawid Lipi{\'n}ski, Agnieszka Kaliska, Paulina Rosalska, Bartosz Topolski, and Przemys{\l}aw Biecek. 2021.
\newblock Kleister: key information extraction datasets involving long documents with complex layouts.
\newblock In \emph{International Conference on Document Analysis and Recognition}, pages 564--579. Springer.

\bibitem[{Tamayo et~al.(2022)Tamayo, Gelbukh, and Burgos}]{tamayo-etal-2022-nlp}
Antonio Tamayo, Alexander Gelbukh, and Diego Burgos. 2022.
\newblock \href {https://aclanthology.org/2022.smm4h-1.6/} {{NLP}-{CIC}-{WFU} at {S}ocial{D}is{NER}: Disease mention extraction in {S}panish tweets using transfer learning and search by propagation}.
\newblock In \emph{Proceedings of The Seventh Workshop on Social Media Mining for Health Applications, Workshop {\&} Shared Task}, pages 19--22, Gyeongju, Republic of Korea. Association for Computational Linguistics.

\bibitem[{Tang et~al.(2023)Tang, Yang, Wang, Fang, Liu, Zhu, Zeng, Zhang, and Bansal}]{tang2023unifying}
Zineng Tang, Ziyi Yang, Guoxin Wang, Yuwei Fang, Yang Liu, Chenguang Zhu, Michael Zeng, Cha Zhang, and Mohit Bansal. 2023.
\newblock Unifying vision, text, and layout for universal document processing.
\newblock In \emph{Proceedings of the IEEE/CVF conference on computer vision and pattern recognition}, pages 19254--19264.

\bibitem[{Team et~al.(2023)Team, Anil, Borgeaud, Alayrac, Yu, Soricut, Schalkwyk, Dai, Hauth, Millican et~al.}]{anil2023gemini}
Gemini Team, Rohan Anil, Sebastian Borgeaud, Jean-Baptiste Alayrac, Jiahui Yu, Radu Soricut, Johan Schalkwyk, Andrew~M Dai, Anja Hauth, Katie Millican, et~al. 2023.
\newblock Gemini: a family of highly capable multimodal models.
\newblock \emph{arXiv preprint arXiv:2312.11805}.

\bibitem[{Van~Landeghem et~al.(2023)Van~Landeghem, Tito, Borchmann, Pietruszka, Joziak, Powalski, Jurkiewicz, Coustaty, Anckaert, Valveny et~al.}]{van2023document}
Jordy Van~Landeghem, Rub{\`e}n Tito, {\L}ukasz Borchmann, Micha{\l} Pietruszka, Pawel Joziak, Rafal Powalski, Dawid Jurkiewicz, Micka{\"e}l Coustaty, Bertrand Anckaert, Ernest Valveny, et~al. 2023.
\newblock Document understanding dataset and evaluation (dude).
\newblock In \emph{Proceedings of the IEEE/CVF International Conference on Computer Vision}, pages 19528--19540.

\bibitem[{Wang et~al.(2023{\natexlab{a}})Wang, Raman, Sibue, Ma, Babkin, Kaur, Pei, Nourbakhsh, and Liu}]{wang2023docllm}
Dongsheng Wang, Natraj Raman, Mathieu Sibue, Zhiqiang Ma, Petr Babkin, Simerjot Kaur, Yulong Pei, Armineh Nourbakhsh, and Xiaomo Liu. 2023{\natexlab{a}}.
\newblock Docllm: A layout-aware generative language model for multimodal document understanding.
\newblock \emph{arXiv preprint arXiv:2401.00908}.

\bibitem[{Wang et~al.(2025)Wang, Zmigrod, Sibue, Pei, Babkin, Brugere, Liu, Navarro, Papadimitriou, Watson, Ma, Nourbakhsh, and Shah}]{zmigrod2024buddiebusinessdocumentdataset}
Dongsheng Wang, Ran Zmigrod, Mathieu~J. Sibue, Yulong Pei, Petr Babkin, Ivan Brugere, Xiaomo Liu, Nacho Navarro, Antony Papadimitriou, William Watson, Zhiqiang Ma, Armineh Nourbakhsh, and Sameena Shah. 2025.
\newblock \href {https://aclanthology.org/2025.finnlp-1.3/} {{B}u{DDIE}: A business document dataset for multi-task information extraction}.
\newblock In \emph{Proceedings of the Joint Workshop of the 9th Financial Technology and Natural Language Processing (FinNLP), the 6th Financial Narrative Processing (FNP), and the 1st Workshop on Large Language Models for Finance and Legal (LLMFinLegal)}, pages 35--47, Abu Dhabi, UAE. Association for Computational Linguistics.

\bibitem[{Wang et~al.(2022)Wang, Chen, Cai, Wei, Yan, Sun, Qin, Li, and Cai}]{j-wang-etal-2022-globalpointer}
Yanbo~J. Wang, Sheng Chen, Hengxing Cai, Wei Wei, Kuo Yan, Zhe Sun, Hui Qin, Yuming Li, and Xiaochen Cai. 2022.
\newblock \href {https://doi.org/10.18653/v1/2022.seretod-1.2} {A {G}lobal{P}ointer based robust approach for information extraction from dialog transcripts}.
\newblock In \emph{Proceedings of the Towards Semi-Supervised and Reinforced Task-Oriented Dialog Systems (SereTOD)}, pages 13--18, Abu Dhabi, Beijing (Hybrid). Association for Computational Linguistics.

\bibitem[{Wang et~al.(2023{\natexlab{b}})Wang, Zhou, Wei, Lee, and Tata}]{Wang_2023}
Zilong Wang, Yichao Zhou, Wei Wei, Chen-Yu Lee, and Sandeep Tata. 2023{\natexlab{b}}.
\newblock \href {https://doi.org/10.1145/3580305.3599929} {Vrdu: A benchmark for visually-rich document understanding}.
\newblock In \emph{Proceedings of the 29th ACM SIGKDD Conference on Knowledge Discovery and Data Mining}, KDD ’23. ACM.

\bibitem[{Wei et~al.(2022)}]{wei2022cot}
Jason Wei et~al. 2022.
\newblock Chain of thought prompting elicits reasoning in large language models.
\newblock \emph{arXiv preprint arXiv:2201.11903}.

\bibitem[{Xu et~al.(2024{\natexlab{a}})Xu, Sun, Zhang, and Zhou}]{xu-etal-2024-chatuie}
Jun Xu, Mengshu Sun, Zhiqiang Zhang, and Jun Zhou. 2024{\natexlab{a}}.
\newblock \href {https://aclanthology.org/2024.lrec-main.279/} {{C}hat{UIE}: Exploring chat-based unified information extraction using large language models}.
\newblock In \emph{Proceedings of the 2024 Joint International Conference on Computational Linguistics, Language Resources and Evaluation (LREC-COLING 2024)}, pages 3146--3152, Torino, Italia. ELRA and ICCL.

\bibitem[{Xu et~al.(2021)Xu, Xu, Lv, Cui, Wei, Wang, Lu, Florencio, Zhang, Che et~al.}]{xu2020layoutlmv2}
Yang Xu, Yiheng Xu, Tengchao Lv, Lei Cui, Furu Wei, Guoxin Wang, Yijuan Lu, Dinei Florencio, Cha Zhang, Wanxiang Che, et~al. 2021.
\newblock Layoutlmv2: Multi-modal pre-training for visually-rich document understanding.
\newblock pages 2579--2591.

\bibitem[{Xu et~al.(2020)Xu, Li, Cui, Huang, Wei, and Zhou}]{xu2020layoutlm}
Yiheng Xu, Minghao Li, Lei Cui, Shaohan Huang, Furu Wei, and Ming Zhou. 2020.
\newblock Layoutlm: Pre-training of text and layout for document image understanding.
\newblock In \emph{Proceedings of the 26th ACM SIGKDD international conference on knowledge discovery \& data mining}, pages 1192--1200.

\bibitem[{Xu et~al.(2024{\natexlab{b}})Xu, Jain, and Kankanhalli}]{xu2024hallucination}
Ziwei Xu, Sanjay Jain, and Mohan Kankanhalli. 2024{\natexlab{b}}.
\newblock Hallucination is inevitable: An innate limitation of large language models.
\newblock \emph{arXiv preprint arXiv:2401.11817}.

\bibitem[{Yang et~al.(2018)Yang, Liu, Liu, Lyu, and Li}]{yang2018adaptations}
An~Yang, Kai Liu, Jing Liu, Yajuan Lyu, and Sujian Li. 2018.
\newblock Adaptations of rouge and bleu to better evaluate machine reading comprehension task.
\newblock In \emph{Proceedings of the Workshop on Machine Reading for Question Answering}, pages 98--104.

\bibitem[{Zhang et~al.(2023)Zhang, Guo, Tu, Chen, Tang, Zhu, Zhang, and Gui}]{zhang-etal-2023-reading}
Chong Zhang, Ya~Guo, Yi~Tu, Huan Chen, Jinyang Tang, Huijia Zhu, Qi~Zhang, and Tao Gui. 2023.
\newblock \href {https://doi.org/10.18653/v1/2023.emnlp-main.846} {Reading order matters: Information extraction from visually-rich documents by token path prediction}.
\newblock In \emph{Proceedings of the 2023 Conference on Empirical Methods in Natural Language Processing}, pages 13716--13730, Singapore. Association for Computational Linguistics.

\bibitem[{Zhang et~al.(2022)Zhang, Ma, Du, Wang, and Zhang}]{zhang2022multimodal}
Zhenrong Zhang, Jiefeng Ma, Jun Du, Licheng Wang, and Jianshu Zhang. 2022.
\newblock Multimodal pre-training based on graph attention network for document understanding.
\newblock \emph{IEEE Transactions on Multimedia}, 25:6743--6755.

\bibitem[{Zhou et~al.(2022)Zhou, Sch{\"a}rli, Hou, Wei, Scales, Wang, Schuurmans, Cui, Bousquet, Le et~al.}]{zhou2022least}
Denny Zhou, Nathanael Sch{\"a}rli, Le~Hou, Jason Wei, Nathan Scales, Xuezhi Wang, Dale Schuurmans, Claire Cui, Olivier Bousquet, Quoc Le, et~al. 2022.
\newblock Least-to-most prompting enables complex reasoning in large language models.
\newblock \emph{arXiv preprint arXiv:2205.10625}.

\bibitem[{Zmigrod et~al.(2024)Zmigrod, Shetty, Sibue, Ma, Nourbakhsh, Liu, and Veloso}]{zmigrod-etal-2024-value}
Ran Zmigrod, Pranav Shetty, Mathieu Sibue, Zhiqiang Ma, Armineh Nourbakhsh, Xiaomo Liu, and Manuela Veloso. 2024.
\newblock \href {https://doi.org/10.18653/v1/2024.findings-emnlp.770} {{\textquotedblleft}what is the value of {templates}?{\textquotedblright} rethinking document information extraction datasets for {LLM}s}.
\newblock In \emph{Findings of the Association for Computational Linguistics: EMNLP 2024}, pages 13162--13185, Miami, Florida, USA. Association for Computational Linguistics.

\end{thebibliography}

\clearpage
\appendix %

\section{Appendix}

\subsection{Discussion \& Observations}
\label{appendix:observations}

\paragraph{\added{Effect of Dataset Characteristics on Parameter Selection.}}
\added{The differences in parameter selection between the VRDU (\textit{OCR--Max--Few-shot--0--Cleaned Pred.--Fuzzy Match}) and FUNSD (\textit{Markdown--Small--Few-shot--0--Cleaned Pred.--Fuzzy Match}) datasets stem from their inherent characteristics. VRDU contains long, multi-page documents extracted via OCR, often resulting in noisy and fragmented text, where using the raw OCR output preserves the original structure. In contrast, FUNSD consists of structured, labeled question-answer pairs, making Markdown a suitable input format for key-value pair extraction. These differences shape the parameter selection process, as some models may inherently favor one format over another due to training data characteristics—e.g., GPT-4o may handle OCR data more effectively, while Llama3 shows a preference for Markdown.}

\paragraph{\added{Unexpected Impact of Markdown Representation on Model Performance.}}
\added{Markdown, while useful for structuring data, does not always improve performance when applied to noisy and unstructured inputs like those from the VRDU dataset. We observed that converting OCR outputs to Markdown could introduce inconsistencies, such as misaligned headers, broken paragraphs, or flattened tables, affecting data representation quality. In contrast, retaining the raw OCR output often preserved the original structure. For structured datasets like FUNSD, however, Markdown aligned well with the data format, facilitating key-value pair extraction. After experimenting with the IE pipeline on VRDU using purely text input without bounding box locations from OCR, we achieved an overall F1 score of 0.632 (with OCR and Markdown results presented in Table 2). These findings demonstrate that (1) LLMs can leverage structural information from both OCR and Markdown, and (2) their performance varies depending on the input format. These results suggest that the effectiveness of Markdown may depend on both dataset characteristics and model preferences, as previously discussed.}

\paragraph{\added{Insights into the Effects of Few-Shot Examples and Chain-of-Thought Prompting.}} \added{Few-shot prompting with zero examples consistently outperforms other configurations due to the model’s ability to generalize without being constrained by specific instances. In VRDU, the variability and noise introduced by OCR, along with the lack of a fixed key-value schema in the FUNSD dataset, make it challenging for the model to generalize from a few specific examples. Few-shot examples often include multiple key-value pairs within a single text that may not appear in other documents, resulting in a mismatch between training and test data. Chain-of-Thought (CoT) prompting, despite its usefulness for reasoning tasks, can overcomplicate straightforward key-value extraction, as guiding the model to think step by step may distract from directly mapping structured content. Moreover, few-shot examples often lack layout information compared to test documents due to the way they were created (see~\ref{appendix:dataset_details}), further contributing to mismatches. Consequently, zero-shot prompting proves more effective by avoiding template-specific biases, maintaining prompt brevity, and enabling better generalization across diverse document structures.}

\paragraph{\added{Potential Optimization Bias in Substring and Fuzzy Matching Evaluations.}} \added{During manual inspection of ground truth and extracted data, we noticed that exact matching often introduces bias by favoring the original annotation, leading to false errors. For example, 'signer\_title' in the ground truth is 'general manager', while the input file and extraction have 'general manager, north america'—both correct, yet exact matching would mark the latter as wrong. Similarly, for 'signer\_name', the ground truth has ['Catherine Redman-randell'], but the extraction includes ['A.j. Maltrentin', 'Catherine Redman-randell'], which is accurate given the context.}

\added{To assess the reliability of substring and fuzzy matching, we manually annotated errors, categorizing them as OCR error, ground truth error, LLM hallucination, additional/wrong information, human error, or incomplete prediction. Substring matching reported zero errors, while fuzzy matching had just one error among 1,730 extractions (see Appendix~\ref{appendix:eval_tech}). These findings support the use of fuzzy matching in scenarios where exact matching would unfairly penalize valid variations.}

\paragraph{\added{Selection Criteria for OFAT and Brute-Force Configurations in Performance Evaluation.}} \added{The best case configurations for each model were determined using a full-factorial (brute-force) exploration, as detailed in Table~\ref{tab:Brute_force_configurations_per_model}, since the optimal configuration for one model may not generalize well to others. In contrast, for the OFAT configurations, we selected the parameter set that achieved the highest average performance across comparisons—specifically, \textit{OCR--Max--Few-shot--0--Cleaned Pred.--Fuzzy Match}—as shown in Tables~\ref{tab:input_type}--\ref{tab:post_processing}.}

\added{While it might seem inconsistent to use OFAT for overall results and brute-force for best/worst cases, the rationale is that OFAT aims to identify generally optimal parameters with minimal effort, whereas the brute-force method is necessary to assess each model’s best and worst performance individually for comparison purposes.}

\subsection{Prompt Generation Details}
\label{appendix:prompt_generation_details}
\begin{figure}[!htb]
\centering
\begin{tcolorbox}[colframe=myyellow, colback=mylightyellow, sharp corners=south, title=Prompt Template: Few-shot,
boxsep=0.5mm,
left=1mm,
right=1mm,
top=1mm,
bottom=1mm]
\footnotesize

\begin{lstlisting}[frame=none, 
                  basicstyle=\ttfamily, 
                  tabsize=4, 
                  morekeywords={Document, Task, Extraction},
                  stringstyle=\color{myblue}]
        ### Examples ###
(<Document>) 
{DOCUMENT_REPRESENTATION} 
(</Document>) 
(<Task>) 
{TASK_DESCRIPTION} 
{SCHEMA_REPRESENTATION} 
(</Task>) 
(<Extraction>) 
{EXTRACTION} 
(</Extraction>)
        ### New Documents ###
(<Document>) 
{DOCUMENT_REPRESENTATION} 
(</Document>) 
(<Task>) 
{TASK_DESCRIPTION} 
{SCHEMA_REPRESENTATION} 
(</Task>) 
(<Extraction>)
\end{lstlisting}
\end{tcolorbox}
\caption{Few-Shot Prompt Structure with 1-shot Example.}
\label{fig:few-shot-prompt}
\rmspace
\end{figure}
\begin{figure}[ht]
\centering
\begin{tcolorbox}[colframe=myyellow, colback=mylightyellow, sharp corners=south, title=Prompt Template: CoT,
boxsep=0.5mm,
left=1mm,
right=1mm,
top=1mm,
bottom=1mm]
\footnotesize
\begin{lstlisting}[frame=none, 
                  basicstyle=\ttfamily, 
                  tabsize=4, 
                  escapeinside={  }, 
                  morekeywords={Document, Task, Extraction, Reasoning},  
                  stringstyle=\color{myblue}]
        ### Examples ###
(<Document>) 
{DOCUMENT_REPRESENTATION} 
(</Document>) 
(<Task>) 
{TASK_DESCRIPTION} 
{SCHEMA_REPRESENTATION} 
(</Task>)
(<Reasoning>) 
{REASONING} 
(</Reasoning>)
(<Extraction>) 
{EXTRACTION} 
(</Extraction>)
        ### New Documents ###
(<Document>) 
{DOCUMENT_REPRESENTATION} 
(</Document>) 
(<Task>) 
{TASK_DESCRIPTION} 
{SCHEMA_REPRESENTATION} 
(</Task>) 
(<Reasoning>) 
{REASONING} 
(</Reasoning>)
(<Extraction>)
\end{lstlisting}
\end{tcolorbox}
\caption{Chain of Thought Prompt Structure with 1-shot Example.}
\label{fig:chain-of-thought-prompt}
\end{figure}

\textit{Document representation} in the example section of the prompt is generated by the LLM, condensing the original OCR data while retaining information relevant to the target schema to reduce token usage and cost. This version integrates only text features. However, in the new example section, both text and layout features are preserved without modification, embedding spatial structure into text sequences to enhance model understanding. This is the only part that differs between the example and new document sections of the prompt. \textit{Task descriptions} provide clear extraction guidelines, specifying what information to retrieve, while \textit{schema representation} defines the expected JSON format to ensure consistency in extracted data. Additionally, CoT prompting includes a \textit{reasoning} component, guiding the model through logical steps to improve accuracy on complex tasks (see Figures~\ref{fig:few-shot-prompt},~\ref{fig:chain-of-thought-prompt}).

\subsection{Tailoring Dataset for our Test-Suite}
\label{appendix:dataset_details}
VRDU dataset includes two benchmarks, each including train samples of 10, 50, 100, and 200 documents. The Registration Form with six entity types used for this project, see Figure~\ref{fig:vrdu_registration_entities}.
\begin{figure}[!htb]
\footnotesize
\begin{lstlisting}[frame=single, basicstyle=\ttfamily, tabsize=4]
{
  "file_date": "",
  "foreign_principle_name": "",
  "registrant_name": "",
  "registration_num": "",
  "signer_name": "",
  "signer_title": ""
}
\end{lstlisting}
\caption{VRDU Registration Form Entities.} 
\label{fig:vrdu_registration_entities}
\rmspace
\end{figure}

\begin{figure}[ht]
    \centering    
\includegraphics[width=0.98\linewidth]{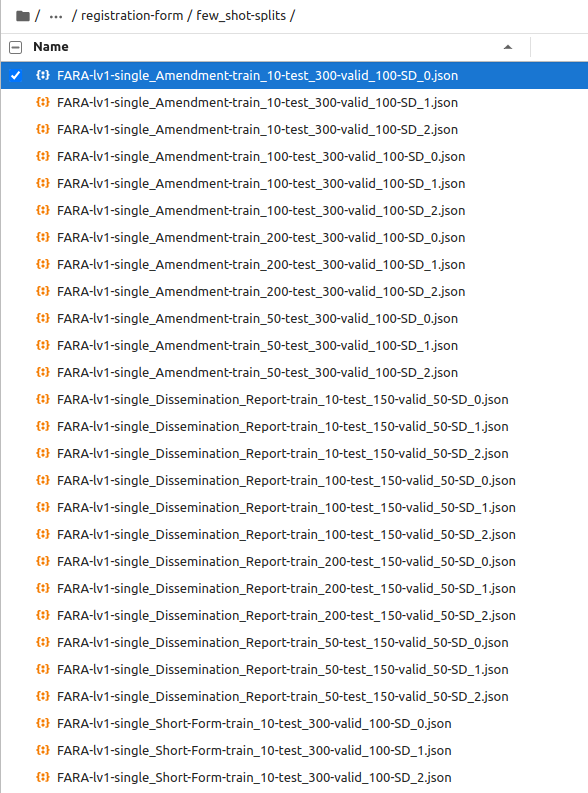}

\caption{Representation of few\_shot\_splits from the VRDU dataset.}
\label{fig:dataset_splits}
\rmspace
\end{figure}

The VRDU dataset includes predefined few-shot splits that consist of train, test, and validation sets. These splits contain 10, 50, 100, and 200 training samples, each with three subsets, as shown in Figure~\ref{fig:dataset_splits}. The dataset also includes different levels (Lv1: Single, Lv2: Mixed, and Lv3: Unseen Type) and various template types (Amendment, Dissemination, and Short-Form).

For each template-level combination, we selected the first JSON file ending in 0 with 10 training samples. Since this training data will be utilized for few-shot and Chain-of-Thought (CoT) prompting, only the first five documents were chosen from the training samples of the selected JSON files. Each level type (STL, UTL) includes template types (Amendment, Dissemination, Short-Form), each with 0, 1, 3, or 5 examples. In STL, these categories use the first document for one-example prompts, the first three for three-example prompts, and all five for five-example prompts. The same structure applies to UTL, with examples specific to its categories. This ensures consistency across template-level combinations while varying the number of examples in the prompt. Figure~\ref{fig:train_examples} shows an example of few-shot and CoT examples. This process was repeated for every level, template type, and example count. The example texts were generated using a Large Language Model (LLM), which was instructed to summarize the provided OCR text for the given document while ensuring the inclusion of target schemas and entities.

\begin{figure}[ht]
\footnotesize
\begin{lstlisting}[frame=single, basicstyle=\ttfamily, tabsize=4]
few_shot_examples = {
  "STL": {
    "Amendment": {
      0: [],
      1: [
        {
          "text": "This document is an amendment to the regis...",
          "entities": {
            "file_date": "1982-10-31",
            "foreign_principle_name": "Japan Trade Center...",
            "registrant_name": "PressAid Center",
            "registration_num": "1833",
            "signer_name": "Akira Tsutsumi",
            "signer_title": "Director General"
          }
        }
      ],
      3: [
        {...},
        {...},
        {...}
      ],
      5: [
        {...},
        {...},
        {...},
        {...},
        {...},
      ]
    },
    "Dissemination": {...
\end{lstlisting}
\caption{VRDU Registration Form Entities.} 
\label{fig:train_examples}
\rmspace
\end{figure}

For the test files, to ensure a fair comparison, we selected the first 40 common files from the chosen JSON files within each template type at each level. This means that for Lv1 Amendment and Lv3 Amendment test files, the first common 40 files were selected as test files, and the same strategy was applied for other template types as well. Due to the mixed nature of test files in Lv2, the mixed template type was excluded from this project. 

For GPT-3.5, GPT-4o, and LLaMA3-70B, we used few-shot and CoT examples, while GPT-4o-vision used the same test dataset without few-shot examples, following basic instructions. For LayoutLMv3 and ERNIE-Layout, we used the same training and test datasets as the other models but included the entire validation set (300 samples). This consistent setup ensured fair evaluation across all models.

\subsection{Success of Evaluation Techniques}
\label{appendix:eval_tech}

We manually reviewed the results of the baseline experiment from GPT-3.5, GPT-4o, and LLaMA3-70B to assess the success of the substring and fuzzy match metrics. The analysis focused on cases where the exact match score was 0 but substring/fuzzy was 1, highlighting predictions that failed strict matching but were successfully handled in other techniques. This created a dataset to test how well substring and fuzzy matching handle difficult cases. In total, we examined 91 key-value pairs for fuzzy match and 37 for substring match.

\begin{figure}[ht]
    \centering
    \includegraphics[width=0.5\textwidth]{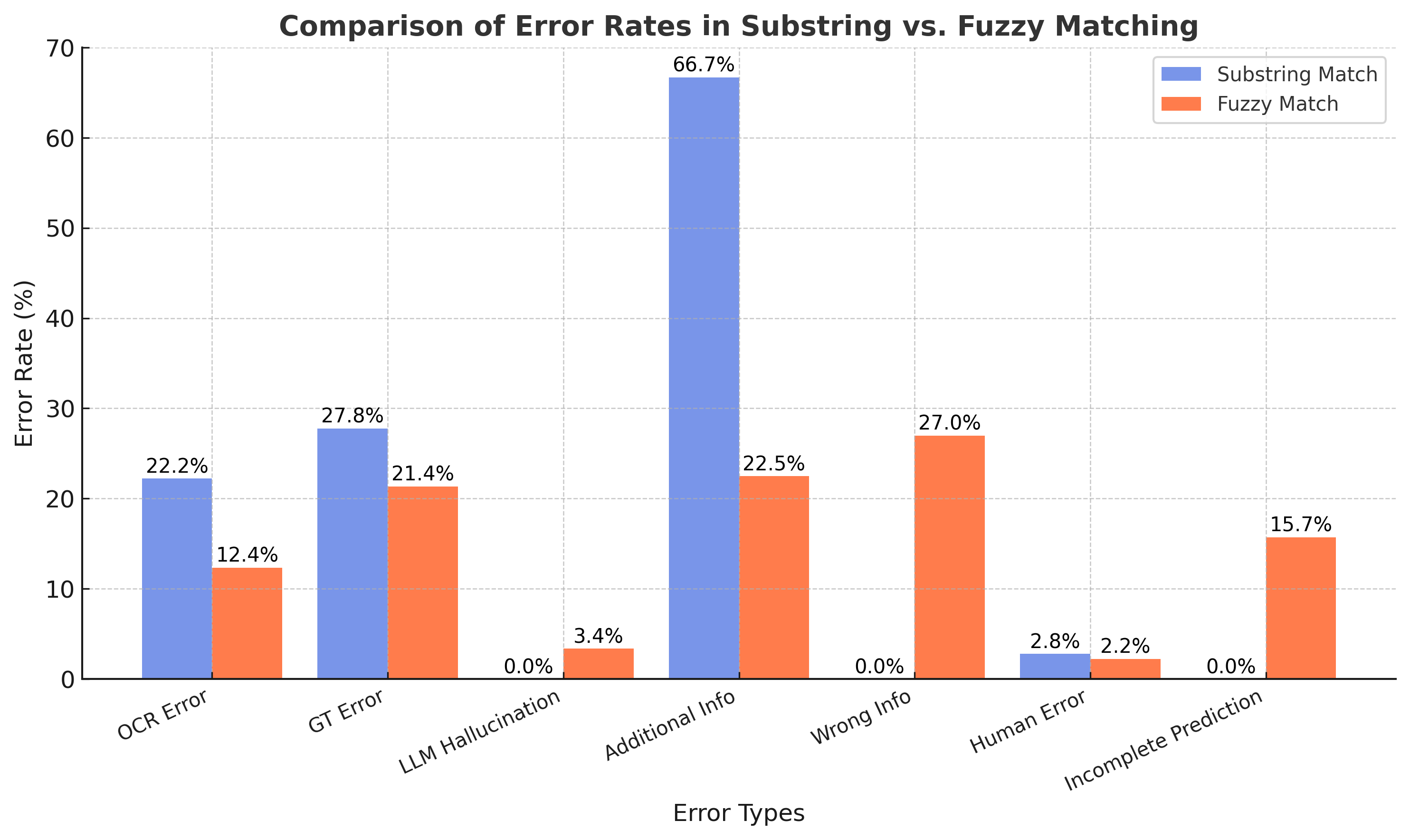}
    \caption{Comparison of error rates in Substring vs. Fuzzy Matching.}
    \label{fig:error_rates}
\rmspace
\end{figure}

Based on our intubation, we identified seven error categories (see Figure~\ref{fig:error_rates}): OCR errors occur when handwriting is misinterpreted (e.g., "Jim Slattery" as "Jim Slatters"). GT errors arise from incomplete or inaccurate ground truth (e.g., "Om Saudi Arabia 1" instead of "Kingdom Saudi Arabia"). LLM hallucinations involve predictions not present in the OCR or document (e.g., predicting "1992-04-24" instead of "1992-04-21"). Additional info errors include correct predictions with extra information (e.g., "Daniel Manatt Todd" instead of "Daniel Manatt"). Wrong info occurs when LLM selects an incorrect value despite the correct one being available (e.g., predicting "2016-10-31" as the file date instead of "2016-10-08").

Human errors result from physical mistakes, such as crooked or incomplete scans. Incomplete predictions occur when the prediction includes part of the ground truth but misses a segment (e.g., "Japan External Trade Organization" instead of "Japan External Trade Organization Tokyo, Japan"). Figure~\ref{fig:error_rates} shows these error types and their rates for both Substring and Fuzzy match categories, where exact matches are labeled as 1.

For evaluating the success of Fuzzy and Substring, we labeled data points as 0 only when the error category is "wrong info"; for other error types, we accepted them and labeled the ground truth as 1. Table~\ref{tab:eval_performances} presents the performance metrics for Fuzzy and Substring calculated based on these ground truth labels, where exact match is 0 and substring/fuzzy is 1.
\begin{table}[hbtp]
    \centering %
    \scriptsize %
    \caption{Performance results of substring and fuzzy evaluation techniques over exact match based on manually annotated data, considering categorized error types.} 
    \label{tab:eval_performances}
    \setlength{\tabcolsep}{4pt}
    \begin{tabular}{l c c c c}
    \toprule
    \textbf{Data} & \textbf{Evaluation} & \multirow{2}{*}{\textbf{Precision}} & \multirow{2}{*}{\textbf{Recall}} & \multirow{2}{*}{\textbf{F1}} \\
    \textbf{Points} & \textbf{Techniques} & & & \\
    \midrule
    \multirow{2}{*}{37} & Exact Match & 0.000 & 0.000 & 0.000 \\
             & Substring Match & 1.000 &  1.000 &  1.000  \\
    \cmidrule(lr){2-5}
    \multirow{2}{*}{91} & Exact Match & 0.000 & 0.000 & 0.000  \\
             & Fuzzy Match & 0.984 & 1.000 & 0.992  \\
    \bottomrule
    \end{tabular}
    \rmspace
\end{table}
\subsection{Additional Experiments with FUNSD Dataset}
\label{appendix:funsd_results}

\added{To demonstrate the generalizability of our techniques, we applied the same steps to the FUNSD dataset as to VRDU.}

\added{For the FUNSD dataset, Markdown input consistently outperforms OCR across all models. GPT-3.5 and LLaMA3 show the most significant improvements (+0.100 and +0.103, respectively), while GPT-4o sees a smaller gain (+0.025). On average, Markdown performs better (0.490 vs. 0.414), with lower variability, indicating its effectiveness for structured data in FUNSD (see Table~\ref{tab:input_type_funsd}).}
\begin{table}[hbtp]
    \centering %
    \small
    \caption{\added{Performance results for different LLMs on the FUNSD dataset with different input types. Baseline configuration in} \colorbox{blue!10}{light blue}.}
    \label{tab:input_type_funsd}
    \setlength{\tabcolsep}{4pt}
    \begin{tabular}{l l l}
        \toprule
        \multirow{2}{*}{\textbf{\added{Models}}} & \multicolumn{2}{c}{\textbf{\added{Exact Match (F1)}}} \\
        \cmidrule(lr){2-3}
        & \textbf{\added{OCR}} & \textbf{\added{Markdown}} \\
        \midrule
        \added{GPT-3.5} & \colorbox{blue!10}{\added{0.357}} & \added{0.457\textbf{{\color{gray}\fontsize{5.5}{8.4}\selectfont(+0.100)}}} \\
        \added{GPT-4o}  & \colorbox{blue!10}{\added{0.498}} & \added{0.523\textbf{{\color{gray}\fontsize{5.5}{8.4}\selectfont(+0.025)}}} \\
        \added{LLaMA3} & \colorbox{blue!10}{\added{0.387}} & \added{0.490\textbf{{\color{gray}\fontsize{5.5}{8.4}\selectfont(+0.103)}}} \\
        \bottomrule
        \textbf{\added{Avg {\color{gray}\fontsize{5.5}{8.4}\selectfont($\pm$ stdev.)}}} 
            & \colorbox{blue!10}{\added{0.414\textbf{{\color{gray}\fontsize{5.5}{8.4}\selectfont($\pm$ 0.074)}}}} 
            & \added{0.490\textbf{{\color{gray}\fontsize{5.5}{8.4}\selectfont($\pm$ 0.033)}}} \\
    \end{tabular}
    \rmspace
\end{table}

\added{Small chunk size yields a slight performance advantage on FUNSD with average F1 of 0.415 over medium (0.414) and max (0.411), likely due to its alignment with FUNSD’s short Q\&A pairs--preserving structure, reducing context dilution, and focusing models on concise, relevant segments (Table~\ref{tab:chunk_size_funsd}).}
\begin{table}[hbtp]
    \small
    \caption{\added{Performance results for different LLMs on the FUNSD dataset with different chunk size categories. Baseline configuration in} \colorbox{blue!10}{light blue}.}
    \label{tab:chunk_size_funsd}
    \resizebox{\linewidth}{!}{
    \setlength{\tabcolsep}{4pt}
    \begin{tabular}{l c c c}
    \toprule
    \multirow{2}{*}{\textbf{\added{Models}}} & \multicolumn{3}{c}{\textbf{\added{Exact Match (F1)}}} \\
    \cmidrule(lr){2-4}
    & \textbf{\added{Small\textbf{{\color{gray}\fontsize{7}{8.4}\selectfont($\leq 1024$)}}}} 
    & \textbf{\added{Medium\textbf{{\color{gray}\fontsize{7}{8.4}\selectfont($\leq 2048$)}}}} 
    & \textbf{\added{Max\textbf{{\color{gray}\fontsize{7}{8.4}\selectfont($\leq 4096$)}}}} \\
    \midrule
    \added{GPT-3.5} & \added{0.354\textbf{{\color{gray}\fontsize{5.5}{8.4}\selectfont(-0.003)}}} 
                                & \colorbox{blue!10}{\added{0.357}} 
                                & \added{0.352\textbf{{\color{gray}\fontsize{5.5}{8.4}\selectfont(-0.005)}}} \\
    \added{GPT-4o}  & \added{0.503\textbf{{\color{gray}\fontsize{5.5}{8.4}\selectfont(+0.005)}}} 
                                & \colorbox{blue!10}{\added{0.498}} 
                                & \added{0.494\textbf{{\color{gray}\fontsize{5.5}{8.4}\selectfont(-0.004)}}} \\
    \added{LLaMA3}  & \added{0.390\textbf{{\color{gray}\fontsize{5.5}{8.4}\selectfont(+0.003)}}} 
                                & \colorbox{blue!10}{\added{0.387}} 
                                & \added{0.389\textbf{{\color{gray}\fontsize{5.5}{8.4}\selectfont(+0.002)}}} \\
    \bottomrule
    \textbf{\added{Avg {\color{gray}\fontsize{5.5}{8.4}\selectfont($\pm$ stdev.)}}} 
        & \added{0.415\textbf{{\color{gray}\fontsize{5.5}{8.4}\selectfont($\pm$ 0.77)}}} 
        & \colorbox{blue!10}{\added{0.414\textbf{{\color{gray}\fontsize{5.5}{8.4}\selectfont($\pm$ 0.074)}}}} 
        & \added{0.411\textbf{{\color{gray}\fontsize{5.5}{8.4}\selectfont($\pm$ 0.073)}}} \\
    \end{tabular}}
    \rmspace
\end{table}

\added{Zero-shot prompting consistently yields the best performance on the FUNSD dataset, with average F1 scores of 0.414 (±0.074) for few-shot prompting and 0.404 (±0.109) for CoT. Few-shot prompting is generally more stable than CoT, especially for GPT-3.5 and GPT-4o. Adding examples usually reduces performance, likely due to noise and limited generalization. LLaMA3 remains more stable with CoT compared to other models, but overall, few-shot prompting is better suited for FUNSD's short, structured Q\&A format (see Table~\ref{tab:prompt_type_with_examples_funsd}).}
\begin{table}[hbtp]
    \centering %
    \small
    \caption{\added{Different LLMs with different prompt types and example numbers. Baseline configuration is highlighted in} \colorbox{blue!10}{light blue}.}
    \label{tab:prompt_type_with_examples_funsd}
    \setlength{\tabcolsep}{1pt}
    \begin{tabular}{l c c c c}
    \toprule
    \multirow{2}{*}{\textbf{\added{Models}}} & \multicolumn{4}{c}{\textbf{\added{Exact Match (F1)}}} \\
    \cmidrule(lr){2-5}
    & \textbf{\added{0}} & \textbf{\added{1}} & \textbf{\added{3}} & \textbf{\added{5}} \\
    \midrule
    \rowcolor{black!10!} \multicolumn{5}{c}{\textbf{\added{\textit{few-shot}}}} \\
    \added{GPT-3.5} & \colorbox{blue!10}{\added{0.357}} & \added{0.289\textbf{{\color{gray}\fontsize{5.5}{8.4}\selectfont(-0.068)}}} & \added{0.283\textbf{{\color{gray}\fontsize{5.5}{8.4}\selectfont(-0.074)}}} & \added{0.316\textbf{{\color{gray}\fontsize{5.5}{8.4}\selectfont(-0.041)}}} \\
    \added{GPT-4o}  & \colorbox{blue!10}{\added{0.498}} & \added{0.467\textbf{{\color{gray}\fontsize{5.5}{8.4}\selectfont(-0.031)}}} & \added{0.399\textbf{{\color{gray}\fontsize{5.5}{8.4}\selectfont(-0.099)}}} & \added{0.417\textbf{{\color{gray}\fontsize{5.5}{8.4}\selectfont(-0.081)}}} \\
    \added{LLaMA3}  & \colorbox{blue!10}{\added{0.387}} & \added{0.397\textbf{{\color{gray}\fontsize{5.5}{8.4}\selectfont(+0.010)}}} & \added{0.398\textbf{{\color{gray}\fontsize{5.5}{8.4}\selectfont(+0.011)}}} & \added{0.393\textbf{{\color{gray}\fontsize{5.5}{8.4}\selectfont(+0.006)}}} \\
    \midrule
    \textbf{\added{Avg {\color{gray}\fontsize{5.5}{8.4}\selectfont($\pm$ stdev.)}}} & 
        \colorbox{blue!10}{\added{0.414\textbf{{\color{gray}\fontsize{5.5}{8.4}\selectfont($\pm$ 0.074)}}}} & 
        \added{0.384\textbf{{\color{gray}\fontsize{5.5}{8.4}\selectfont($\pm$ 0.089)}}} & 
        \added{0.360\textbf{{\color{gray}\fontsize{5.5}{8.4}\selectfont($\pm$ 0.66)}}} & 
        \added{0.375\textbf{{\color{gray}\fontsize{5.5}{8.4}\selectfont($\pm$ 0.052)}}} \\
    \rowcolor{black!10!} \multicolumn{5}{c}{\textbf{\added{\textit{CoT}}}} \\
    \added{GPT-3.5} & \added{0.327\textbf{{\color{gray}\fontsize{5.5}{8.4}\selectfont(-0.030)}}} & \added{0.184\textbf{{\color{gray}\fontsize{5.5}{8.4}\selectfont(-0.173)}}} & \added{0.205\textbf{{\color{gray}\fontsize{5.5}{8.4}\selectfont(-0.152)}}} & \added{0.218\textbf{{\color{gray}\fontsize{5.5}{8.4}\selectfont(-0.139)}}} \\
    \added{GPT-4o}  & \added{0.482\textbf{{\color{gray}\fontsize{5.5}{8.4}\selectfont(-0.016)}}} & \added{0.396\textbf{{\color{gray}\fontsize{5.5}{8.4}\selectfont(-0.102)}}} & \added{0.369\textbf{{\color{gray}\fontsize{5.5}{8.4}\selectfont(-0.129)}}} & \added{0.358\textbf{{\color{gray}\fontsize{5.5}{8.4}\selectfont(-0.140)}}} \\
    \added{LLaMA3}  & \added{0.407\textbf{{\color{gray}\fontsize{5.5}{8.4}\selectfont(+0.020)}}} & \added{0.377\textbf{{\color{gray}\fontsize{5.5}{8.4}\selectfont(-0.10)}}} & \added{0.360\textbf{{\color{gray}\fontsize{5.5}{8.4}\selectfont(-0.027)}}} & \added{0.385\textbf{{\color{gray}\fontsize{5.5}{8.4}\selectfont(-0.002)}}} \\
    \midrule
    \textbf{\added{Avg {\color{gray}\fontsize{5.5}{8.4}\selectfont($\pm$ stdev.)}}} & 
        \added{0.404\textbf{{\color{gray}\fontsize{5.5}{8.4}\selectfont($\pm$ 0.109)}}} & 
        \added{0.290\textbf{{\color{gray}\fontsize{5.5}{8.4}\selectfont($\pm$ 0.149)}}} & 
        \added{0.287\textbf{{\color{gray}\fontsize{5.5}{8.4}\selectfont($\pm$ 0.115)}}} & 
        \added{0.288\textbf{{\color{gray}\fontsize{5.5}{8.4}\selectfont($\pm$ 0.98)}}} \\
    \bottomrule
    \end{tabular}
    \rmspace
\end{table}

\added{The results indicate that Cleaned Prediction consistently yields better or comparable F1 scores compared to Initial Prediction, with minor improvements due to effective noise reduction, while Mapped Prediction consistently reduces performance due to overgeneralization and key misalignment. This issue arises because the FUNSD dataset lacks a fixed schema, causing key-value pairs to vary dynamically between documents, leading to mapping errors (see Table~\ref{tab:post_processing_funsd}).}

\begin{table}[hbtp]
\small
\caption{\added{Different LLMs with different post-processing strategies. Baseline configuration is highlighted in} \colorbox{blue!10}{light blue}.}
\label{tab:post_processing_funsd}
\resizebox{\linewidth}{!}{
\setlength{\tabcolsep}{4pt}
\begin{tabular}{l c c c}
\toprule
\multirow{2}{*}{\textbf{\added{Models}}} & \multicolumn{3}{c}{\textbf{\added{Exact Match (F1)}}} \\
\cmidrule(lr){2-4}
& \textbf{\added{Initial Pred.}} & \textbf{\added{Mapped Pred.}} & \textbf{\added{Cleaned Pred.}} \\
\midrule
\added{GPT-3.5} & \colorbox{blue!10}{\added{0.357}} & \added{0.311\textbf{{\color{gray}\fontsize{5.5}{8.4}\selectfont(-0.046)}}} & \added{0.361\textbf{{\color{gray}\fontsize{5.5}{8.4}\selectfont(+0.004)}}} \\
\added{GPT-4o}  & \colorbox{blue!10}{\added{0.498}} & \added{0.456\textbf{{\color{gray}\fontsize{5.5}{8.4}\selectfont(-0.042)}}} & \added{0.496\textbf{{\color{gray}\fontsize{5.5}{8.4}\selectfont(-0.002)}}} \\
\added{LLaMA3}  & \colorbox{blue!10}{\added{0.387}} & \added{0.348\textbf{{\color{gray}\fontsize{5.5}{8.4}\selectfont(-0.039)}}} & \added{0.393\textbf{{\color{gray}\fontsize{5.5}{8.4}\selectfont(+0.006)}}} \\
\bottomrule
\textbf{\added{Avg {\color{gray}\fontsize{5.5}{8.4}\selectfont($\pm$ stdev.)}}} 
& \colorbox{blue!10}{\added{0.414\textbf{{\color{gray}\fontsize{5.5}{8.4}\selectfont($\pm$ 0.074)}}}} 
& \added{0.371\textbf{{\color{gray}\fontsize{5.5}{8.4}\selectfont($\pm$ 0.075)}}} 
& \added{0.416\textbf{{\color{gray}\fontsize{5.5}{8.4}\selectfont($\pm$ 0.070)}}} \\
\end{tabular}}
\rmspace
\end{table}

\added{The results indicate that Substring Match and Fuzzy Match consistently outperform Exact Match across all models, with average F1 scores of 0.488 (±0.060) and 0.497 (±0.092) respectively, compared to 0.414 (±0.074) for Exact Match. This improvement is due to the more flexible matching criteria of substring and fuzzy evaluation, which better accommodate minor variations or partial matches in predictions. The performance gain is most pronounced in GPT-4o, highlighting that models producing slightly varied outputs benefit significantly from more tolerant evaluation metrics (see Table~\ref{tab:evaluation_techniques_funsd}).}
\begin{table}[hbtp]
    \small
    \caption{\added{Different LLMs with different evaluation techniques. Baseline configuration is highlighted in} \colorbox{blue!10}{light blue}.}
    \label{tab:evaluation_techniques_funsd}
    \resizebox{\linewidth}{!}{
    \setlength{\tabcolsep}{4pt}
    \begin{tabular}{l c c c}
    \toprule
    \multirow{2}{*}{\textbf{\added{Models}}} & \multicolumn{3}{c}{\textbf{\added{Evaluation Metric (F1)}}} \\
    \cmidrule(lr){2-4}
    & \textbf{\added{Exact Match}} & \textbf{\added{Substring Match}} & \textbf{\added{Fuzzy Match}} \\
    \midrule
    \added{GPT-3.5} & \colorbox{blue!10}{\added{0.357}} & \added{0.440\textbf{{\color{gray}\fontsize{5.5}{8.4}\selectfont(+0.083)}}} & \added{0.423\textbf{{\color{gray}\fontsize{5.5}{8.4}\selectfont(+0.066)}}} \\
    \added{GPT-4o}  & \colorbox{blue!10}{\added{0.498}} & \added{0.557\textbf{{\color{gray}\fontsize{5.5}{8.4}\selectfont(+0.059)}}} & \added{0.601\textbf{{\color{gray}\fontsize{5.5}{8.4}\selectfont(+0.103)}}} \\
    \added{LLaMA3}  & \colorbox{blue!10}{\added{0.387}} & \added{0.469\textbf{{\color{gray}\fontsize{5.5}{8.4}\selectfont(+0.082)}}} & \added{0.468\textbf{{\color{gray}\fontsize{5.5}{8.4}\selectfont(+0.081)}}} \\
    \bottomrule
    \textbf{\added{Avg {\color{gray}\fontsize{5.5}{8.4}\selectfont($\pm$ stdev.)}}} 
        & \colorbox{blue!10}{\added{0.414\textbf{{\color{gray}\fontsize{5.5}{8.4}\selectfont($\pm$ 0.074)}}}} 
        & \added{0.488\textbf{{\color{gray}\fontsize{5.5}{8.4}\selectfont($\pm$ 0.060)}}} 
        & \added{0.497\textbf{{\color{gray}\fontsize{5.5}{8.4}\selectfont($\pm$ 0.092)}}} \\
    \end{tabular}}
    \rmspace
\end{table}

\added{As OFAT, drawing from the underlying 12 experiments, we identified the optimal parameter for each step and each model based on the experimental outcomes (Table~\ref{tab:OFAT_configurations_per_model_funsd}). Also, we conducted an exhaustive full factorial exploration with 432 configurations to find the best parametrization per LLM (Table~\ref{tab:Brute_force_configurations_per_model_funsd}). Lastly, we find the worst configuration on a per-LLM and per-model basis.}

\begin{table}[hbtp]
\centering
\footnotesize
\caption{\added{OFAT configurations on a per-model basis and corresponding performance results.}}
\begin{tabular}{lcccc}
\toprule
\textbf{\added{Parameter}} & \textbf{\added{GPT-3.5}} & \textbf{\added{GPT-4o}} & \textbf{\added{LLaMA3}} \\
\midrule
\added{Input Type} & \added{Markdown} & \added{Markdown} & \added{Markdown}  \\
\added{Chunk Size} & \added{Medium} & \added{Small} & \added{Small} \\
\added{Prompt} & \added{Few-Shot} & \added{Few-Shot} & \added{CoT} \\
\added{Example No.} & \added{0} & \added{0} & \added{0} \\
\added{Output Refin.} & \added{Cleaned} & \added{Initial} & \added{Cleaned} \\
\added{Evaluation} & \added{Substring} & \added{Fuzzy} & \added{Substring} \\
\bottomrule
\end{tabular}
\rmspace
\label{tab:OFAT_configurations_per_model_funsd}
\end{table}

\begin{table}[hbtp]
\centering
\footnotesize
\caption{\added{Brute-Force configurations on a per-model basis and corresponding performance results.}}
\begin{tabular}{lcccc}
\toprule
\textbf{\added{Parameter}} & \textbf{\added{GPT-3.5}} & \textbf{\added{GPT-4o}} & \textbf{\added{LLaMA3}} \\
\midrule
\added{Input Type} & \added{Markdown} & \added{Markdown} & \added{Markdown}  \\
\added{Chunk Size} & \added{Max} & \added{Medium} & \added{Small} \\
\added{Prompt} & \added{Few-shot} & \added{CoT} & \added{Few-shot} \\
\added{Example No.} & \added{3} & \added{0} & \added{0} \\
\added{Output Refin.} & \added{Cleaned} & \added{Cleaned} & \added{Cleaned} \\
\added{Evaluation} & \added{Fuzzy} & \added{Fuzzy} & \added{Fuzzy} \\
\bottomrule
\end{tabular}
\rmspace
\label{tab:Brute_force_configurations_per_model_funsd}
\end{table}

\begin{figure}[ht]
    \centering
    \includegraphics[width=\columnwidth]{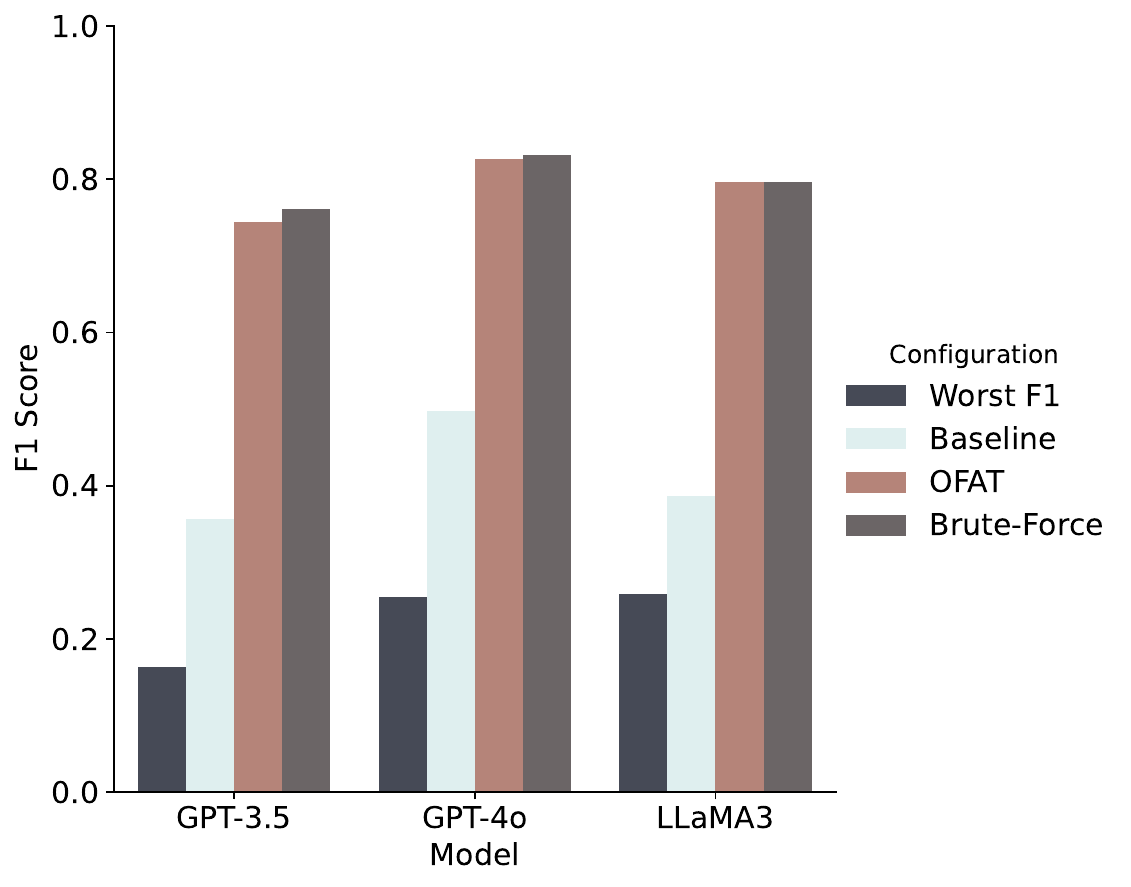}
    \caption{\added{F1 scores of different LLMs across three configurations (Baseline, OFAT, and Brute-Force). Each bar represents the mean F1 score of a model for the corresponding configuration.\final{The OFAT configuration uses a single global parameter set optimized across all models.}}}
    \label{fig:configuration_comparison_funsd}
\end{figure}

\added{The performance of these different configurations is depicted in Figure~\ref{fig:configuration_comparison_funsd}. The OFAT method achieves an average F1 score of 0.789, marking a substantial gain of $37.5$ points over the baseline configuration (0.414). In comparison, the full factorial exploration yields a slightly higher average F1 score of 0.797--only $0.8$ points more--while evaluating all 432 configurations. Notably, OFAT achieves this performance by exploring just 12 configurations, requiring approximately 2.8\% of the computational effort compared to the full factorial search space. Thus, OFAT captures over 99\% of the brute-force method’s effectiveness, making it a highly efficient alternative when computational resources are limited.
Overall, OFAT continues to offer a compelling balance between performance and efficiency. In comparison, the worst configuration achieves only 0.226 on average, which is approximately 28\% of the best configuration}.

\end{document}